\documentclass[fleqn,10pt]{olplainarticle}
\usepackage{algorithm,algpseudocode, subcaption, hyperref}
\usepackage[flushleft]{threeparttable}

\usepackage{varwidth}
\DeclareCaptionFormat{varwidth}{%
  \begin{varwidth}{\linewidth}#1#2#3\end{varwidth}%
}
\makeatletter
\renewcommand\paragraph{\@startsection{paragraph}{4}{\z@}%
                                     {-3.25ex\@plus -1ex \@minus -.2ex}%
                                     {1.5ex \@plus .2ex}%
                                     {\normalfont\normalsize\bfseries}}
\makeatother

\title{Boost Event-Driven Tactile Learning with Location Spiking Neurons}

\author{Peng Kang$^{1}$, Srutarshi Banerjee$^{2}$, Henry Chopp$^{2}$, Aggelos Katsaggelos$^{2}$, and Oliver Cossairt}
\affil[1]{Department of Computer Science, Northwestern, Evanston, IL, USA}
\affil[2]{Department of Electrical and Computer Engineering, Northwestern, Evanston, IL, USA}


\begin{abstract}
Tactile sensing is essential for a variety of daily tasks. Inspired by the event-driven nature and sparse spiking communication of the biological systems, recent advances in event-driven tactile sensors and Spiking Neural Networks (SNNs) spur the research in related fields. However, SNN-enabled event-driven tactile learning is still in its infancy due to the limited representation abilities of existing spiking neurons and high spatio-temporal complexity in the event-driven tactile data. In this paper, to improve the representation capability of existing spiking neurons, we propose a novel neuron model called ``location spiking neuron'',  which enables us to extract features of event-based data in a novel way. Specifically, based on the classical \underline{T}ime \underline{S}pike \underline{R}esponse \underline{M}odel (TSRM), we develop the \underline{L}ocation \underline{S}pike \underline{R}esponse \underline{M}odel (LSRM). In addition, based on the most commonly-used \underline{T}ime \underline{L}eaky \underline{I}ntegrate-and-\underline{F}ire (TLIF) model, we develop the \underline{L}ocation \underline{L}eaky \underline{I}ntegrate-and-\underline{F}ire (LLIF) model\footnote[1]{TSRM is the classical \underline{S}pike \underline{R}esponse \underline{M}odel (SRM) in the literature and TLIF is the classical \underline{L}eaky \underline{I}ntegrate-and-\underline{F}ire (LIF) in the literature. We add the character ``T'' to highlight their differences from LSRM and LLIF.}. Moreover, to demonstrate the representation effectiveness of our proposed neurons and capture the complex spatio-temporal dependencies in the event-driven tactile data, we exploit the location spiking neurons to propose two hybrid models for event-driven tactile learning. Specifically, the first hybrid model combines a fully-connected SNN with TSRM neurons and a fully-connected SNN with LSRM neurons. And the second hybrid model fuses the spatial spiking graph neural network with TLIF neurons and the temporal spiking graph neural network with LLIF neurons. Extensive experiments demonstrate the significant improvements of our models over the state-of-the-art methods on event-driven tactile learning, including event-driven tactile object recognition and event-driven slip detection. Moreover, compared to the counterpart artificial neural networks (ANNs), our SNN models are \textbf{10$\times$} to \textbf{100$\times$} energy-efficient, which shows the superior energy efficiency of our models and may bring new opportunities to the spike-based learning community and neuromorphic engineering. 
\end{abstract}

\begin{document}

\flushbottom
\maketitle
\thispagestyle{empty}

\section{Introduction}\label{intro}

With the prevalence of artificial intelligence, computers today have demonstrated extraordinary abilities in visual and auditory perceptions. Although these perceptions are essential sensory modalities, they may fail to complete tasks in certain situations where tactile perception can help. For example, the visual sensory modality can fail to distinguish objects with similar visual features in less-favorable environments, such as dim-lit or in the presence of occlusions. In such cases, tactile sensing can provide meaningful information like texture, pressure, roughness, or friction and maintain performance. Overall, tactile perception is a vital sensing modality that enables humans to gain perceptual judgment on the surrounding environment and conduct stable movements~\cite{taunyazov2020fast}. 

With the recent advances in material science and Artificial Neural Networks (ANNs), research on tactile perception has begun to soar, including tactile object recognition~\cite{soh2014incrementally, kappassov2015tactile, sanchez2018online}, slip detection~\cite{calandra2018more}, and texture recognition~\cite{baishya2016robust, taunyazov2019towards}. Unfortunately, although ANNs demonstrate promising performance on the tactile learning tasks, they are usually power-hungry compared to human brains that require far less energy to perform the tactile perception robustly~\cite{li2016evaluating, strubell2019energy}. 

Inspired by biological systems, research on event-driven perception has started to gain momentum, and several asynchronous event-based sensors have been proposed, including event cameras~\cite{gallego2020event} and event-based tactile sensors~\cite{taunyazov2020event}. In contrast to standard synchronous sensors, such event-based sensors can achieve higher energy efficiency, better scalability, and lower latency. However, due to the high sparsity and complexity of event-driven data, learning with these sensors is still in its infancy~\cite{pfeiffer2018deep}. Recently, several works~\cite{taunyazov2020event, gu2020tactilesgnet, taunyazov2020fast} utilized Spiking Neural Networks (SNNs)~\cite{Shrestha2018, pfeiffer2018deep, ijcai2021-441} to tackle event-driven tactile learning. Unlike ANNs, which require expensive transformations from asynchronous discrete events to synchronous real-valued frames, SNNs can process event-based sensor data directly. Moreover, unlike ANNs that employ artificial neurons~\cite{maas2013rectifier, xu2015empirical, clevert2015fast} and conduct real-valued computations, SNNs adopt spiking neurons~\cite{gerstner1995time, abbott1999lapicque, gerstner2002spiking} and utilize binary 0-1 spikes to process information. This difference reduces the mathematical dot-product operations in ANNs to less computationally expensive summation operations in SNNs~\cite{roy2019towards}. Due to the advantages of SNNs, these works are always energy-efficient and suitable for power-constrained devices. However, due to the limited representative abilities of existing spiking neuron models and high spatio-temporal complexity in the event-based tactile data~\cite{taunyazov2020event}, these works still cannot sufficiently capture spatio-temporal dependencies and thus hinder the performance of event-driven tactile learning.

In this paper, to address the problems mentioned above, we make several contributions that boost event-driven tactile learning, including event-driven tactile object recognition and event-driven slip detection. We summarize a list of acronyms and notations in Table~\ref{notation_tab}. Please refer to it during the reading.

\textbf{First, to enable richer representative abilities of existing spiking neurons, we propose a novel neuron model called ``location spiking neuron''.} Unlike existing spiking neuron models that update their membrane potentials based on time steps~\cite{roy2019towards}, location spiking neurons update their membrane potentials based on locations. Specifically, based on the Time Spike Response Model (TSRM)~\cite{gerstner1995time}, we develop the ``Location Spike Response Model (LSRM)''. Moreover, to make the location spiking neurons more applicable to a wide range of applications, we develop the ``Location Leaky Integrate-and-Fire (LLIF)'' model based on the most commonly-used Time Leaky Integrate-and-Fire (TLIF) model~\cite{abbott1999lapicque}. Please note that TSRM  and TLIF are the classical Spike Response Model (SRM) and Leaky Integrate-and-Fire (LIF) in the literature. We add the character ``T (Time)'' to highlight their differences from LSRM and LLIF. These location spiking neurons enable the extraction of feature representations of event-based data in a novel way. Previously, SNNs adopted temporal recurrent neuronal dynamics to extract features from the event-based data. With location spiking neurons, we can build SNNs that employ spatial recurrent neuronal dynamics to extract features from the event-based data. We believe location spiking neuron models can have a broad impact on the SNN community and spur the research on spike-based learning from event sensors like NeuTouch~\cite{taunyazov2020event}, Dynamic Audio Sensors~\cite{anumula2018feature}, or Dynamic Vision Sensors~\cite{gallego2020event}.

\textbf{Next, we investigate the representation effectiveness of location spiking neurons and propose two models for event-driven tactile learning.} Specifically, to capture the complex spatio-temporal dependencies in the event-driven tactile data, the first model combines a fully-connected (FC) SNN with TSRM neurons and a fully-connected (FC) SNN with LSRM neurons, henceforth referred to as the \textbf{Hybrid\_SRM\_FC}. To capture more spatio-temporal topology knowledge in the event-driven tactile data, the second model fuses the spatial spiking graph neural network (GNN) with TLIF neurons and temporal spiking graph neural network (GNN) with LLIF neurons, henceforth referred to as the \textbf{Hybrid\_LIF\_GNN}. To be more specific, the Hybrid\_LIF\_GNN first constructs tactile spatial graphs and tactile temporal graphs based on taxel locations and event time sequences, respectively. Then, it utilizes the spatial spiking graph neural network with TLIF neurons and the temporal spiking graph neural network with LLIF neurons to extract features of these graphs. Finally, it fuses the spiking tactile features from the two networks and provides the final tactile learning prediction. Besides the novel model construction, we also specify the location orders to enable the spatial recurrent neuronal dynamics of location spiking neurons in event-driven tactile learning. In addition, we explore the robustness of location orders on event-driven tactile learning. Moreover, we design new loss functions involved with locations and utilize the backpropagation methods to optimize the proposed models. Furthermore, we develop the timestep-wise inference algorithms for the two models to show their applicability to the spike-based temporal data.


%

\textbf{Lastly, we conduct experiments on three challenging event-driven tactile learning tasks.} Specifically, the first task requires models to determine the type of objects being handled. The second task requires models to determine the type of containers being handled and the amount of liquid held within, which is more challenging than the first task. And the third task asks models to accurately detect the rotational slip (``stable'' or ``rotate'') within 0.15s. Extensive experimental results demonstrate the significant improvements of our models over the state-of-the-art methods on event-driven tactile learning. Moreover, the experiments show that existing spiking neurons are better at capturing spatial dependencies, while location spiking neurons are better at modeling mid-and-long temporal dependencies. Furthermore, compared to the counterpart ANNs, our models are \textbf{10$\times$} to \textbf{100$\times$} energy-efficient, which shows the superior energy efficiency of our models and may bring new opportunities to neuromorphic engineering. 


Portions of this work ``Event-Driven Tactile Learning with Location Spiking Neurons~\cite{kangTactile}'' were accepted by IJCNN 2022 and an oral presentation was given at the IEEE WCCI 2022. We highlight the additional contributions in this paper.
\begin{itemize}
	\item To make the location spiking neurons user-friendly in various spike-based learning frameworks, we expand the idea of location spiking neurons to the most commonly-used TLIF neurons and propose the LLIF neurons. Specifically, the LLIF neurons update their membrane potentials based on locations and enable the models to extract features with spatial recurrent neuronal dynamics. We can incorporate the LLIF neurons into popular spike-based learning frameworks like STBP~\cite{wu2018spatio} and tap their feature representation potential. We believe such neuron models can have a broad impact on the SNN community and spur the research on spike-based learning.
	\item To demonstrate the advantage of LLIF neurons and further boost the event-based tactile learning performance, we build the Hybrid\_LIF\_GNN, which fuses the spatial spiking graph neural network with TLIF neurons and the temporal spiking graph neural network with LLIF neurons. The model extracts features from tactile spatial graphs and tactile temporal graphs concurrently. To the best of our knowledge, this is the first work to construct tactile temporal graphs based on event sequences and build a temporal spiking graph neural network for event-driven tactile learning.
	\item We further include more data,  experiments, and interpretation to demonstrate the effectiveness and energy efficiency of the proposed neurons and models. Extensive experiments on real-world datasets show that the Hybrid\_LIF\_GNN significantly outperforms the state-of-the-art methods for event-driven tactile learning, including the Hybrid\_SRM\_FC~\cite{kangTactile}. Moreover, the computational cost evaluation demonstrates the high-efficiency benefits of the Hybrid\_LIF\_GNN and LLIF neurons, which may unlock their potential on neuromorphic hardware. The source code is available at \url{https://github.com/pkang2017/TactileLSN}.
	\item We thoroughly discuss the advantages and limitations of existing spiking neurons and location spiking neurons. Moreover, we provide preliminary results on event-driven audio learning and discuss the broad applicability and potential impact of this work on other spike-based learning applications.
\end{itemize}

The rest of the paper is organized as follows. In Section~\ref{sec:related}, we provide an overview of related work on SNNs and event-driven tactile sensing and learning. In Section~\ref{sec:methods}, we start by introducing notations for existing spiking neurons and extend them to the specific location spiking neurons. We then propose various models with location spiking neurons for event-driven tactile learning. Last, we provide implementation details and algorithms related to the proposed models. In Section~\ref{sec:exp}, we demonstrate the effectiveness and energy efficiency of our models on benchmark datasets. Finally, we discuss and conclude in Section~\ref{sec:conclude}.

\section{Related Work}\label{sec:related}
In the following, we provide a brief overview of related work on SNNs and event-driven tactile sensing and learning.
\subsection{Spiking Neural Networks (SNNs)}
With the prevalence of Artificial Neural Networks (ANNs), computers today have demonstrated extraordinary abilities in many cognition tasks. However, ANNs only imitate brain structures in several ways, including the vast connectivity and structural and functional organizational hierarchy \cite{roy2019towards}. The brain has more information processing mechanisms like the neuronal and synaptic functionality \cite{bullmore2012economy, felleman1991distributed}. Moreover, ANNs are much more energy-consuming than human brains. To integrate more brain-like characteristics and make artificial intelligence models more energy-efficient, researchers propose Spiking Neural Networks (SNNs), which can be executed on power-efficient neuromorphic processors like TrueNorth~\cite{merolla2014million} and Loihi~\cite{davies2021advancing}. Similar to ANNs, SNNs can adopt general network topologies like convolutional layers and fully-connected layers, but use different neuron models~\cite{gerstner2002spiking}, such as the Time Leaky Integrate-and-Fire (TLIF) model~\cite{abbott1999lapicque} and the Time Spike Response Model (TSRM)~\cite{gerstner1995time}. Due to the non-differentiability of these spiking neuron models, it still remains challenging to train SNNs. Nevertheless, several solutions have been proposed, such as converting the trained ANNs to SNNs~\cite{cao2015spiking,sengupta2019going} and approximating the derivative of the spike function~\cite{wu2018spatio, ijcai2020-211}. In this work, we propose location spiking neurons to enhance the representative abilities of existing spiking neurons. These location spiking neurons maintain the spiking characteristic but employ the spatial recurrent neuronal dynamics, which enable us to build energy-efficient SNNs and extract features of event-based data in a novel way. Moreover, based on the optimization methods for SNNs with existing spiking neurons, we design new loss functions for SNNs with location spiking neurons and utilize the backpropagation methods with surrogate gradients to optimize the proposed models.
\subsection{Event-Driven Tactile Sensing and Learning}
With the prevalence of material science and robotics, several tactile sensors have been developed, including non-event-based tactile sensors like the iCub RoboSkin~\cite{schmitz2010tactile} and the SynTouch BioTac\cite{fishel2012sensing} and event-driven tactile sensors like the NeuTouch~\cite{taunyazov2020event} and the NUSkin~\cite{taunyazov2021extended}. In this paper, we focus on event-driven tactile learning with SNNs. Since the development of event-driven tactile sensors is still in its infancy~\cite{gu2020tactilesgnet}, little prior work exists on learning event-based tactile data with SNNs. The work~\cite{taunyazov2020fast} employed a neural coding scheme to convert raw tactile data from non-event-based tactile sensors into event-based spike trains. It then utilized an SNN to process the spike trains and classify textures. A recent work~\cite{taunyazov2020event} released the first publicly-available event-driven visual-tactile dataset collected by NeuTouch and proposed an SNN based on SLAYER~\cite{Shrestha2018} to solve the event-driven tactile learning. Moreover, to naturally capture the spatial topological relations and structural knowledge in the event-based tactile data, a very recent work~\cite{gu2020tactilesgnet} utilized the spiking graph neural network~\cite{ijcai2021-441} to process the event-based tactile data and conduct the tactile object recognition. In this paper, different from previous works building SNNs with spiking neurons that employ the temporal recurrent neuronal dynamics, we construct SNNs with location spiking neurons to capture the complex spatio-temporal dependencies in the event-based tactile data and improve event-driven tactile learning.



\section{Methods}\label{sec:methods}

In this section, we first demonstrate the spatial recurrent neuronal dynamics of location spiking neurons by introducing notations for the existing spiking neurons and extending them to the location spiking neurons. We then introduce two models with location spiking neurons for event-driven tactile learning. Last, we provide implementation details and algorithms related to the proposed models.

\begin{figure}[ht]
     \centering
     \begin{subfigure}[b]{0.3\textwidth}
         \centering
         \includegraphics[width=\textwidth]{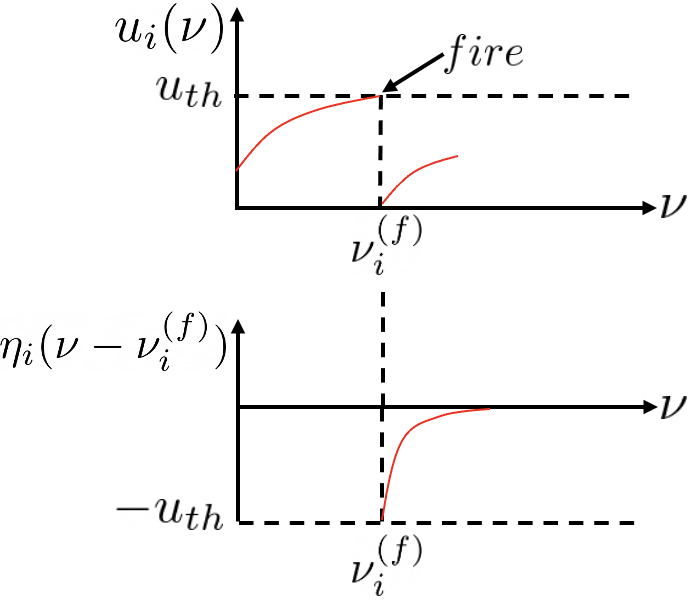}
         \caption{}
         \label{fig:dynamics1}
     \end{subfigure}
     \hfill
     \begin{subfigure}[b]{0.3\textwidth}
         \centering
         \includegraphics[width=\textwidth]{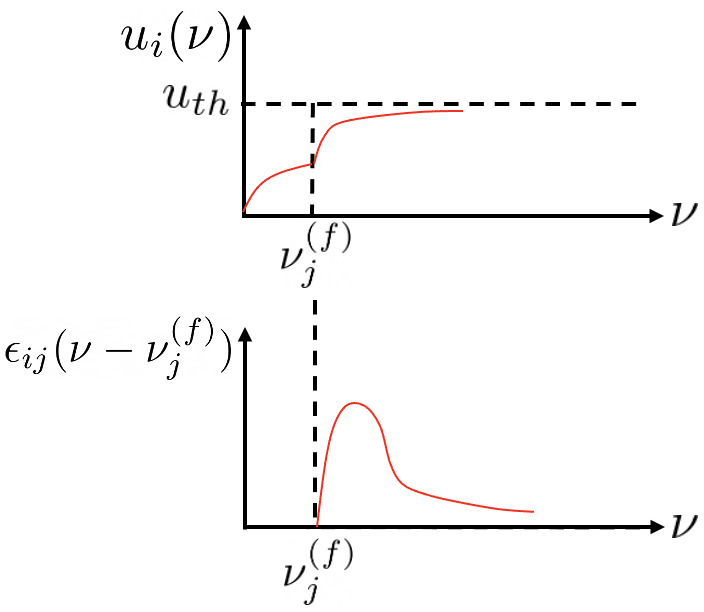}
         \caption{}
         \label{fig:dynamics2}
     \end{subfigure}
     \hfill
     \begin{subfigure}[b]{0.3\textwidth}
         \centering
         \includegraphics[width=\textwidth]{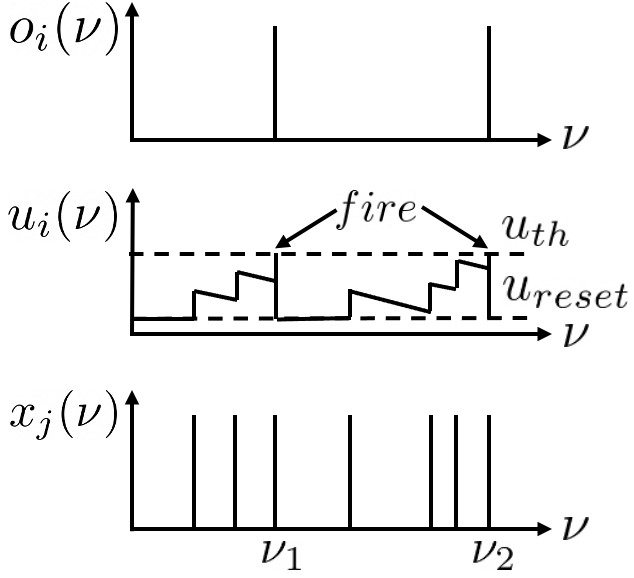}
         \caption{}
         \label{fig:dynamics3}
     \end{subfigure}
     \caption{Recurrent neuronal dynamic mechanisms for the existing spiking neurons of $\nu=t$ and location spiking neurons of $\nu=l$. \textbf{(a)} The refractory dynamics of a TSRM neuron $i$ or an LSRM neuron $i$. Immediately after firing an output spike at $\nu_i^{(f)}$, the value of $u_i(\nu)$ is lowered or reset by adding a negative contribution $\eta_i(\cdot)$. The kernel $\eta_i(\cdot)$ vanishes for $\nu<\nu_i^{(f)}$ and decays to zero for $\nu\to\infty$. \textbf{(b)} The incoming spike dynamics of a TSRM neuron $i$ or an LSRM neuron $i$. A presynaptic spike at $\nu_j^{(f)}$ increases the value of $u_i(\nu)$ for $\nu\geq \nu_j^{(f)}$ by an amount of $w_{ij}x_j(\nu_j^{(f)})\epsilon_{ij}(\nu-\nu_j^{(f)})$. The kernel $\epsilon_{ij}(\cdot)$ vanishes for $\nu<\nu_j^{(f)}$. ``$<$'' and ``$\geq$'' indicate the location order when $\nu=l$. \textbf{(c)} The recurrent neuronal dynamics of a TLIF neuron $i$ or an LLIF neuron $i$. The neuron $i$ takes as input binary spikes and outputs binary spikes. $x_j$ represents the input signal to the neuron $i$ from neuron $j$, $u_i$ is the neuron's membrane potential, and $o_i$ is the neuron's output. An output spike will be emitted from the neuron when its membrane potential surpasses the firing threshold $u_{th}$, after which the membrane potential will be reset to $u_{reset}$.}
     \label{fig:dynamics}
\end{figure}

\subsection{Existing Spiking Neuron Models vs. Location Spiking Neuron Models}
Spiking neuron models are mathematical descriptions of specific cells in the nervous system. They are the basic building blocks of SNNs. In this section, we first introduce the mechanisms of existing spiking neuron models -- the TSRM and the TLIF. To enrich their representative abilities, we transform them into location spiking neuron models -- the LSRM and the LLIF. 


In the TSRM, the temporal recurrent neuronal dynamics of neuron $i$ are described by its membrane potential $u_i(t)$. When $u_i(t)$ exceeds a predefined threshold $u_{th}$ at the firing time $t_i^{(f)}$, the neuron $i$ will generate a spike. The set of all firing times of neuron $i$ is denoted by 
\begin{equation}
	\label{e1}
	\mathcal{F}_i=\{t_i^{(f)}; 1\leq f\leq n\} = \{t|u_i(t) = u_{th}\}, 
\end{equation}  
where $t_i^{(n)}$ is the most recent spike time $t_i^{(f)}<t$. The value of $u_i(t)$ is governed by two different spike response processes:
\begin{equation}
	\label{e2}
	u_i(t) = \sum_{t_i^{(f)}\in\mathcal{F}_i}\eta_i(t-t_i^{(f)}) + \sum_{j\in\Gamma_i}\sum_{t_j^{(f)}\in\mathcal{F}_j}w_{ij}x_j(t_j^{(f)})\epsilon_{ij}(t-t_j^{(f)}), 
\end{equation}  
where $\Gamma_i$ is the set of presynaptic neurons of neuron $i$ and $x_j(t_j^{(f)})=1$ is the presynaptic spike at time $t_j^{(f)}$. $\eta_i(t)$ is the refractory kernel, which describes the response of neuron $i$ to its own spikes at time $t$. $\epsilon_{ij}(t)$ is the incoming spike response kernel, which models the neuron $i$'s response to the presynaptic spikes from neuron $j$ at time $t$. $w_{ij}$ accounts for the connection strength between neuron $i$ and neuron $j$ and scales the incoming spike response. Figure~\ref{fig:dynamics1} of $\nu=t$ visualizes the refractory dynamics of the TSRM neuron $i$ and Figure~\ref{fig:dynamics2} of $\nu=t$ visualizes the incoming spike dynamics of the TSRM neuron $i$. 

Without loss of generality, such temporal recurrent neuronal dynamics also apply to other spiking neuron models, such as the TLIF, which is a special case of the TSRM~\cite{maass2001pulsed}. Since the TLIF model is computationally tractable and maintains biological fidelity to a certain degree, it becomes the most commonly-used spiking neuron model and there are many popular SNN frameworks powered with it~\cite{wu2018spatio}. The dynamics of the TLIF neuron $i$ are governed by
\begin{equation}
	\label{e5}
	\tau \frac{du_i(t)}{dt} = -u_i(t) + I(t), 
\end{equation}
where $u_i(t)$ represents the internal membrane potential of the neuron $i$ at time $t$, $\tau$ is a time constant, and $I(t)$ signifies the presynaptic input obtained by the combined action of synaptic weights and pre-neuronal activities. To better understand the membrane potential update of TLIF neurons, the Euler method is used to transform the first-order differential equation of Eq.~(\ref{e5}) into a recursive expression:
\begin{equation}
	\label{e6}
	u_i(t) = (1-\frac{dt}{\tau})u_i(t-1) + \frac{dt}{\tau}\sum_jw_{ij}x_j(t),
\end{equation}
where $\sum_jw_{ij}x_j(t)$ is the weighted summation of the inputs from pre-neurons at the current time step. Equation~(\ref{e6}) can be further simplified as:
\begin{equation}
	\label{e7}
	u_i(t) = \alpha u_i(t-1) + \sum_jw_{ij}'x_j(t),
\end{equation}
where $\alpha=1-\frac{dt}{\tau}$ can be considered a decay factor, and $w_{ij}'$ is the weight incorporating the scaling effect of $\frac{dt}{\tau}$. When $u_i(t)$ exceeds a certain threshold $u_{th}$, the neuron emits a spike, resets its membrane potential to $u_{reset}$, and then accumulates $u_i(t)$ again in subsequent time steps. Figure~\ref{fig:dynamics3} of $\nu=t$ visualizes the temporal dynamics of a TLIF neuron $i$. 

From the above descriptions, we find that existing spiking neuron models have \textbf{explicit temporal recurrence} but do not possess \textbf{explicit spatial recurrence}, which, to some extent, limits their representative abilities. 
To enrich the representative abilities of existing spiking neuron models, we propose location spiking neurons, which adopt the spatial recurrent neuronal dynamics and update their membrane potentials based on locations\footnote{locations could refer to pixel or patch locations for images or taxel locations for tactile sensors.}. These neurons exploit \textbf{explicit spatial recurrence}. Specifically, the spatial recurrent neuronal dynamics of the LSRM neuron $i$ are described by its location membrane potential $u_i(l)$. When $u_i(l)$ exceeds a predefined threshold $u_{th}$ at the firing location $l_i^{(f)}$, the neuron $i$ will generate a spike. The set of all firing locations of neuron $i$ is denoted by 
\begin{equation}
	\label{e3}
	\mathcal{G}_i=\{l_i^{(f)}; 1\leq f\leq n\} = \{l|u_i(l) = u_{th}\}, 
\end{equation}  
where $l_i^{(n)}$ is the nearest firing location $l_i^{(f)} < l$. ``$<$'' indicates the location order, which is manually set and will be discussed in Section~\ref{alg}. The value of $u_i(l)$ is governed by two different spike response processes:
\begin{equation}
	\label{e4}
	u_i(l) = \sum_{l_i^{(f)}\in\mathcal{G}_i}\eta_i(l-l_i^{(f)}) + \sum_{j\in\Gamma_i}\sum_{l_j^{(f)}\in\mathcal{G}_j}w_{ij}x_j(l_j^{(f)})\epsilon_{ij}(l-l_j^{(f)}), 
\end{equation}  
where $\Gamma_i$ is the set of presynaptic neurons of neuron $i$ and $x_j(l_j^{(f)})=1$ is the presynaptic spike at location $l_j^{(f)}$. $\eta_i(l)$ is the refractory kernel, which describes the response of neuron $i$ to its own spikes at location $l$. $\epsilon_{ij}(l)$ is the incoming spike response kernel, which models the neuron $i$'s response to the presynaptic spikes from neuron $j$ at location $l$. Figure~\ref{fig:dynamics1} of $\nu=l$ visualizes the refractory dynamics of the LSRM neuron $i$ and Figure~\ref{fig:dynamics2} of $\nu=l$ visualizes the incoming spike dynamics of the LSRM neuron $i$. The threshold $u_{th}$ of LSRM neurons can be different from that of TSRM neurons, while we set the same for simplicity. In Section~\ref{Tactile-LSRM}, we will apply the LSRM neurons to event-driven tactile learning and show how the proposed neurons enable feature extraction in a novel way.

To make the location spiking neurons user-friendly and compatible with various spike-based learning frameworks, we expand the idea of location spiking neurons to the most commonly-used TLIF neurons and propose the LLIF neurons. Different from the temporal dynamics shown in Eq.~(\ref{e5}), the LLIF neuron $i$ employs the spatial dynamics:
\begin{equation}
	\label{e8}
	\tau' \frac{du_i(l)}{dl} = -u_i(l) + I(l),
\end{equation}
where $u_i(l)$ represents the internal membrane potential of an LLIF neuron $i$ at location $l$, $\tau'$ is a location constant, and $I(l)$ represents the presynaptic input. We use the Euler method again to transform the first-order differential equation of Eq.~(\ref{e8}) into a recursive expression:
\begin{align}
	\label{e9}
	u_i(l) = (1-\frac{dl}{\tau'})u_i(l_{prev}) + \frac{dl}{\tau'}\sum_jw_{ij}x_j(l),
\end{align}
where $\sum_jw_{ij}x_j(l)$ is the weighted summation of the inputs from pre-neurons at the current location. Equation~(\ref{e9}) can be further simplified as:
\begin{equation}
	\label{e10}
	u_i(l) = \beta u_i(l_{prev}) + \sum_jw_{ij}'x_j(l),
\end{equation}
where $\beta=1-\frac{dl}{\tau'}$ can be considered a location decay factor, and $w_{ij}'$ is the weight incorporating the scaling effect of $\frac{dl}{\tau'}$. When $u_i(l)$ exceeds a certain threshold $u_{th}$, the neuron emits a spike, resets its membrane potential to $u_{reset}$, and then accumulates $u_i(l)$ again at subsequent locations. $u_{th}$ and $u_{reset}$ of LLIF neurons can be different from those of TLIF neurons, while we set the same for simplicity. Figure~\ref{fig:dynamics3} of $\nu=l$ visualizes the spatial recurrent neuronal dynamics of an LLIF neuron $i$. To enable the dynamics of LLIF neurons, we still need to specify the location order like the LSRM neurons. In Section~\ref{Tactile-LLIF}, we will demonstrate how the LLIF neurons can be incorporated into the popular spike-based learning framework and further boost the performance of event-driven tactile learning.
\begin{figure*}
	\includegraphics[width=\linewidth]{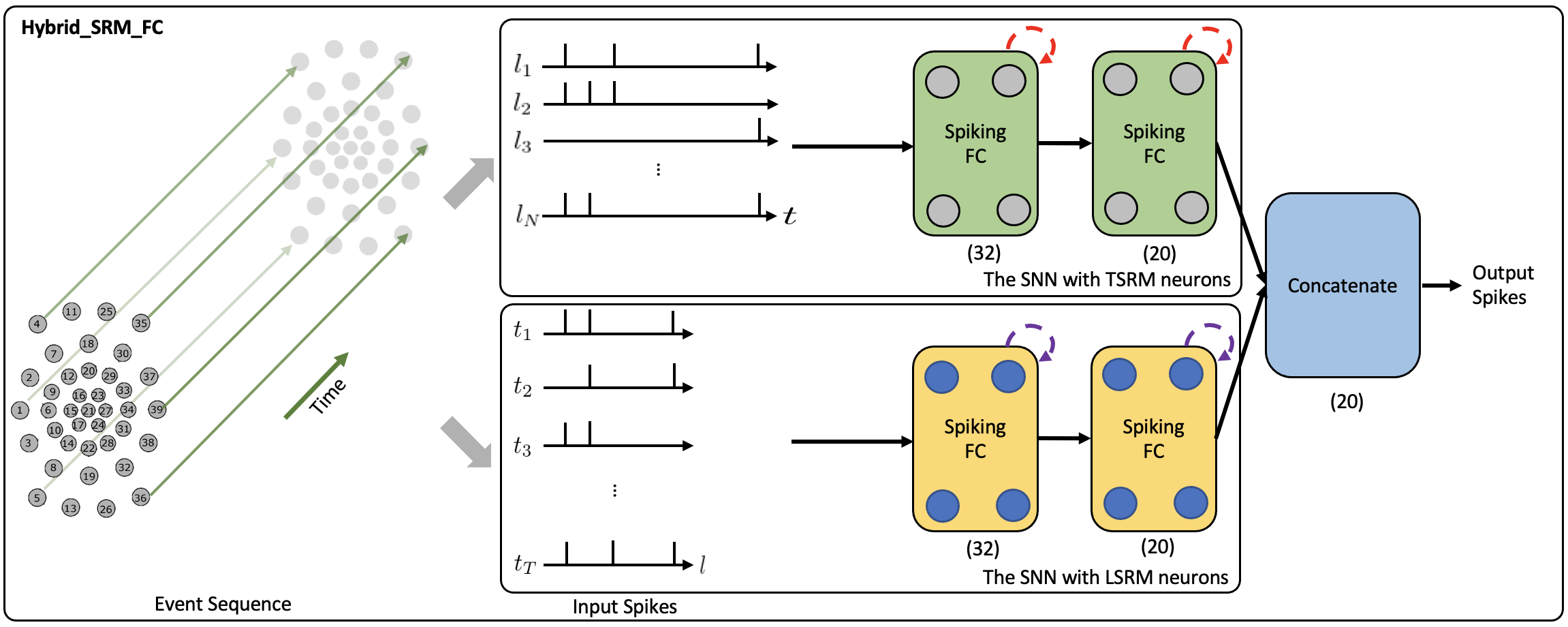}
	\caption{The network structure of the Hybrid\_SRM\_FC. The SNN with TSRM neurons processes the input spikes $X_{in}$ and adopts the temporal recurrent neuronal dynamics (shown with red dashed arrows) of TSRM neurons to extract features from the data. The SNN with LSRM neurons processes the transposed input spikes $X'_{in}$ and employs the spatial recurrent neuronal dynamics (shown with purple dashed arrows) of LSRM neurons to extract features from the data. Finally, the spiking representations from two networks are concatenated to yield the final predicted label. (32) and (20) represent the sizes of fully-connected layers, where we assume the number of classes ($K$) is 20. This figure is adapted from~\cite{kangTactile}.}
	\label{hybrid}
\end{figure*}
\subsection{Event-Driven Tactile Learning with Location Spiking Neurons}
To investigate the representation effectiveness of location spiking neurons and boost the event-driven tactile learning performance, we propose two models with location spiking neurons, which capture complex spatio-temporal dependencies in the event-based tactile data. In this paper, we focus on processing the data collected by NeuTouch~\cite{taunyazov2020event}, a biologically-inspired event-driven fingertip tactile sensor with 39 taxels arranged spatially in a radial fashion (see Fig.~\ref{hybrid}). 

\subsubsection{Event-Driven Tactile Learning with the LSRM Neurons}\label{Tactile-LSRM}
In this section, we introduce event-driven tactile learning with the LSRM neurons. Specifically, we propose the Hybrid\_SRM\_FC to capture the complex spatio-temporal dependencies in the event-driven tactile data.

Figure~\ref{hybrid} presents the network structure of the Hybrid\_SRM\_FC. From the figure, we can see that the model has two components, including the fully-connected SNN with TSRM neurons and the fully-connected SNN with LSRM neurons. Specifically, the fully-connected SNN with TSRM neurons employs the temporal recurrent neuronal dynamics to extract spiking feature representations from the event-based tactile data $X_{in} \in \mathbb{R}^{N\times T}$, where $N$ is the total number of taxels and $T$ is the total time length of event sequences. The fully-connected SNN with LSRM neurons utilizes the spatial recurrent neuronal dynamics to extract spiking feature representations from the event-based tactile data $X'_{in} \in \mathbb{R}^{T\times N}$, where $X'_{in}$ is transposed from $X_{in}$. The spiking representations from two networks are then concatenated to yield the final task-specific output.

To be more specific, the top part of Fig.~\ref{hybrid} shows the network structure of fully-connected SNN with TSRM neurons. It employs two spiking fully-connected layers with TSRM neurons to process $X_{in}$ and generate the spiking representations $O_{1}\in\mathbb{R}^{K\times T}$, where $K$ is the output dimension determined by the task. The membrane potential $u_i(t)$, the output spiking state $o_i(t)$, and the set of all firing times $\mathcal{F}_i $ of TSRM neuron $i$ in these layers are decided by:
\begin{equation}	\label{e11}
	\begin{split}
		u_i(t) &= \sum_{t_i^{(f)}\in\mathcal{F}_i}\eta(t-t_i^{(f)}) + \underbrace{\sum_{j\in\Gamma_i}\sum_{t_j^{(f)}\in\mathcal{F}_j}w_{ij}o_j(t_j^{(f)})\epsilon(t-t_j^{(f)})}_\textbf{capture spatial dependencies},   \\
		o_i(t) &= \begin{cases}
			1 & \text{if $u_i(t) \geq u_{th}$};\\
			0 & \text{otherwise},
		\end{cases}  \\
		\mathcal{F}_i &= \begin{cases}
			\mathcal{F}_i \cup t & \text{if $o_i(t) = 1$};\\
			\mathcal{F}_i & \text{otherwise},
		\end{cases}           
	\end{split}
\end{equation}
where $w_{ij}$ are the trainable parameters, $\eta$($t$) and $\epsilon$($t$) model the temporal recurrent neuronal dynamics of TSRM neurons, \textbf{$\Gamma_i$ is the set of presynaptic TSRM neurons spanning over the spatial domain, which is utilized to capture the spatial dependencies in the event-based tactile data.}

Moreover, the bottom part of Fig.~\ref{hybrid} shows the network structure of fully-connected SNN with LSRM neurons. It employs two spiking fully-connected layers with LSRM neurons to process $X'_{in}$ and generate the spiking representations $O_{2}\in\mathbb{R}^{K\times N}$, where $K$ is the output dimension decided by the task. The membrane potential $u_i(l)$, the output spiking state $o_i(l)$, and the set of all firing locations $\mathcal{G}_i $ of LSRM neuron $i$ in these layers are decided by:
\begin{equation}	\label{e12}
	\begin{split}
		u_i(l) &= \sum_{l_i^{(f)}\in\mathcal{G}_i}\eta(l-l_i^{(f)}) + \underbrace{\sum_{j\in\Gamma'_i}\sum_{l_j^{(f)}\in\mathcal{G}_j}w_{ij}o_j(l_j^{(f)})\epsilon(l-l_j^{(f)})}_\textbf{model temporal dependencies},   \\
		o_i(l) &= \begin{cases}
			1 & \text{if $u_i(l) \geq u_{th}$};\\
			0 & \text{otherwise},
		\end{cases}   \\
		\mathcal{G}_i &= \begin{cases}
			\mathcal{G}_i \cup l & \text{if $o_i(l) = 1$};\\
			\mathcal{G}_i & \text{otherwise},
		\end{cases}           
	\end{split}    
\end{equation}
where $w_{ij}$ are the trainable connection weights, $\eta$($l$) and $\epsilon$($l$) determine the spatial recurrent neuronal dynamics of LSRM neurons, \textbf{$\Gamma'_i$ is the set of presynaptic LSRM neurons spanning over the temporal domain, which is utilized to model the temporal dependencies in the event-based tactile data.} Such location spiking neurons tap the representative potential and enable us to capture features in this novel way.

Lastly, we concatenate the spiking representations of $O_{1}$ and $O_{2}$ along the last dimension and obtain the final output spike train $O\in\mathbb{R}^{K\times (T+N)}$. The predicted label is associated with the neuron $k\in K$ with the largest number of spikes in the domain of $T+N$. 

\subsubsection{Event-Driven Tactile Learning with the LLIF Neurons}\label{Tactile-LLIF}
In this section, to demonstrate the usability of location spiking neurons and further boost the event-driven tactile learning performance, we utilize the LLIF neurons to propose the Hybrid\_LIF\_GNN, which fuses spatial and temporal spiking graph neural networks and captures complex spatio-temporal dependencies in the event-based tactile data. 

\paragraph{Tactile Graph Construction}

\begin{figure}
     \centering
     \begin{subfigure}[b]{0.49\textwidth}
         \centering
         \includegraphics[height=75mm, width=\textwidth]{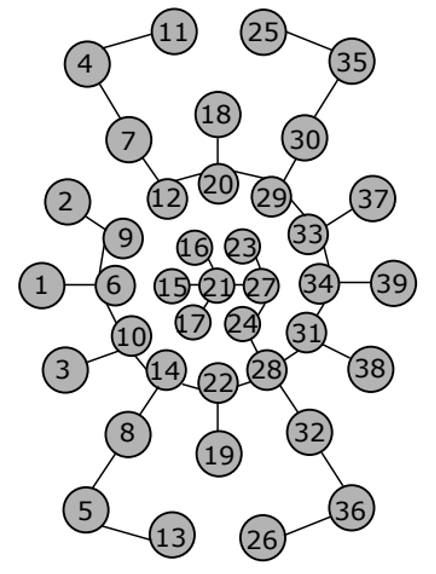}
         \caption{}
         \label{fig:graph1}
     \end{subfigure}
     \hfill
     \begin{subfigure}[b]{0.49\textwidth}
         \centering
         \includegraphics[width=\textwidth]{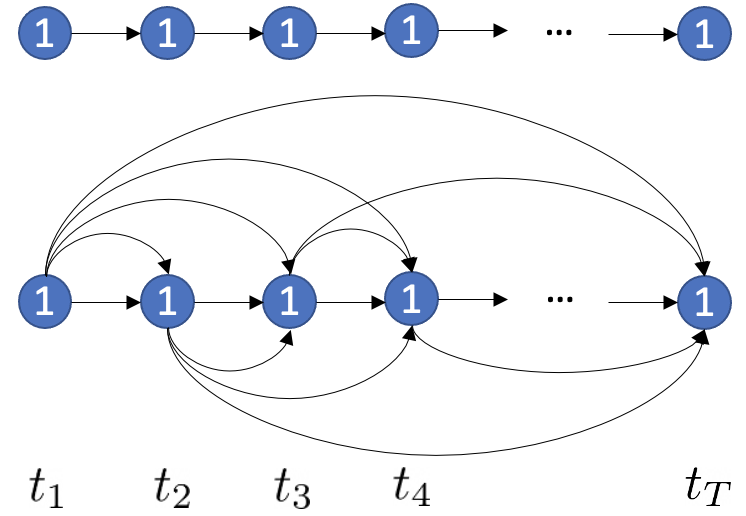}
         \caption{}
         \label{fig:graph2}
     \end{subfigure}
        \caption{\textbf{(a)} The tactile spatial graph $G_s$ at time step $t$ generated by the Minimum Spanning Tree (MST) algorithm~\cite{gu2020tactilesgnet}. Each circle represents a taxel of NeuTouch. \textbf{(b)} Based on event sequences, we propose two different tactile temporal graphs $G_t$ for a specific taxel $n=1$: the above one is the sparse tactile temporal graph, while the below one is the dense tactile temporal graph.}
        \label{graphs}
\end{figure}




Given event-based tactile inputs $X_{in} \in \mathbb{R}^{N\times T}$, we construct tactile spatial graphs and tactile temporal graphs as illustrated in Fig.~\ref{graphs}. 


The tactile spatial graph $G_{s}(t) = (V^{t}, E^{t})$ at time step $t$ \textbf{explicitly captures the spatial structural information in the data}, while the tactile temporal graph $G_{t}(n) = (V_{n}, E_{n})$ for a specific taxel $n$ \textbf{explicitly models the temporal dependency in the data}. $V^{t} = \{v^t_n|n=1,...,N\}$ and $V_{n} = \{v^t_n|t=1,...,T\}$ represent nodes of $G_s(t)$ and $G_t(n)$, respectively, and the attribute of $v^t_n$ is the event feature of the $n$-th taxel at time step $t$. $E^t = \{e^t_{i,j}|i,j=1,...,N\}$ represents the edges of $G_s(t)$, where $e^t_{i,j}\in \{0,1\}$ indicates whether the nodes $v^t_i$, $v^t_j$ are connected (denoted as 1) or disconnected (denoted as 0). $E^t$ is formed by the Minimum Spanning Tree (MST) algorithm, where the Euclidean distance between taxels $d(v^t_i, v^t_j) = \|(x, y)_{v^t_i} - (x, y)_{v^t_j}\|_2$ is used to determine whether the edges are in the MST. Since the 2D coordinates $(x,y)$ of taxels do not change with time, $E^t$ remains the same throughout time. Moreover, the adjacency matrix of $E^t$ is symmetric (i.e., the edges are indirect) as we assume the mutual spatial dependency in the data. $E_{n} = \{e_n^{p,q}|p,q=1,...,T\}$ represents the edges of $G_{t}(n)$, where $e_n^{p,q}\in \{0,1\}$ and each edge is direct. Based on different temporal dependency assumptions, we propose two kinds of tactile temporal graphs shown in Fig.~\ref{fig:graph2}. One is sparse since we assume the current state only directly impacts the nearest future state. While the other is dense since we assume the current state has a broad impact on the future states. $E_{n}$ remains the same for all $N$ taxels.


\begin{figure*}[ht]
	\includegraphics[width=\linewidth]{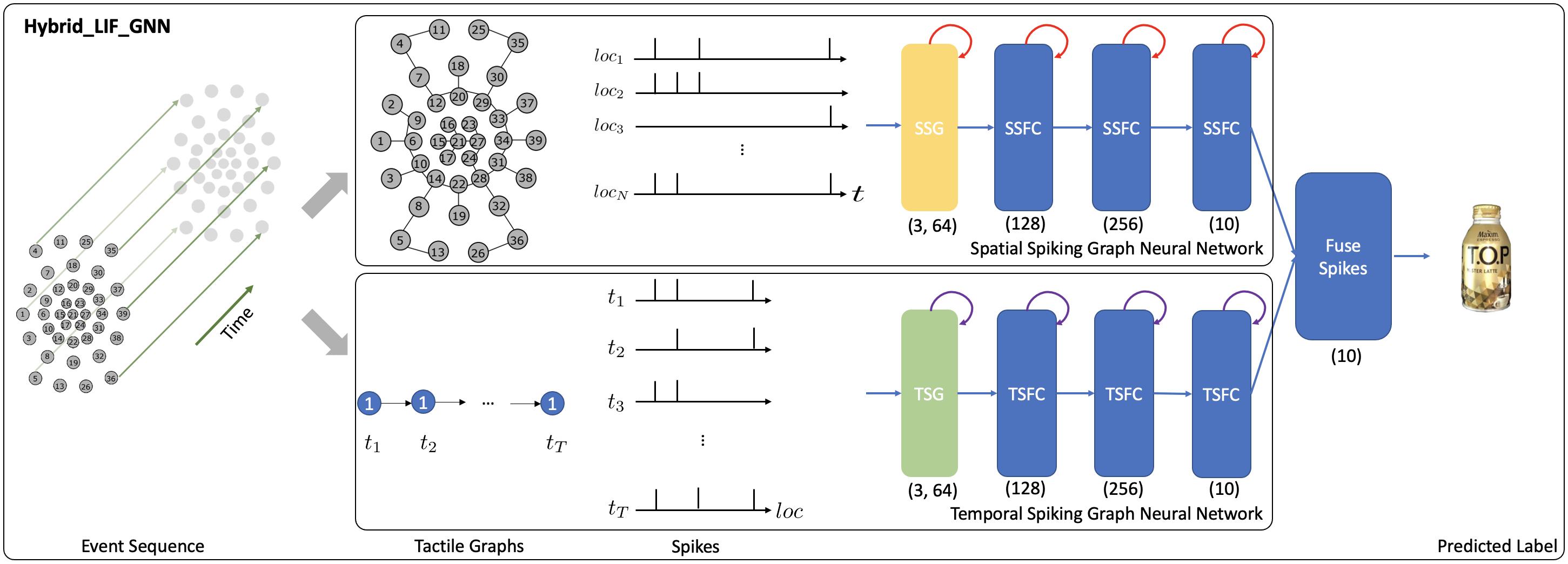}
	\caption{The structure of the Hybrid\_LIF\_GNN, where ``SSG'' is the spatial spiking graph layer, ``SSFC'' is the spatial spiking fully-connected layer, ``TSG'' is the temporal spiking graph layer, and ``TSFC'' is the temporal spiking fully-connected layer. The spatial spiking graph neural network processes the $T$ tactile spatial graphs and adopts the temporal recurrent neuronal dynamics (shown with red arrows) of TLIF neurons to extract features. The temporal spiking graph neural network processes the $N$ tactile temporal graphs and employs the spatial recurrent neuronal dynamics (shown with purple arrows) of LLIF neurons to extract features. Finally, the model fuses the predictions from two networks and obtains the final predicted label. (3, 64) represents the hop size and the filter size of spiking graph layers. (128), (256), and (10) represent the sizes of fully-connected layers, where we assume the number of classes ($K$) is 10.}
	\label{structures}
\end{figure*}

\paragraph{Hybrid\_LIF\_GNN}
To process the data from tactile graphs and capture the complex spatio-temporal dependencies in the event-based tactile data, we propose the Hybrid\_LIF\_GNN (see Fig.~\ref{structures}), which fuses spatial and temporal spiking graph neural networks. Specifically, we adopt the spatial spiking graph neural network with TLIF neurons~\cite{gu2020tactilesgnet}, which is a spike-based tactile learning framework powered by STBP~\cite{wu2018spatio}. It uses temporal recurrent neuronal dynamics to capture the spatial structure information from the tactile spatial graphs. Inspired by this model, we develop the temporal spiking graph neural network with LLIF neurons, which is also powered by STBP. Our temporal spiking graph neural network utilizes spatial recurrent neuronal dynamics to extract the temporal dependencies in the tactile temporal graphs. Finally, we fuse the spiking features from two networks and obtain the final prediction. 

To be more specific, the spatial spiking graph neural network takes as input tactile spatial graphs, and it has one spatial spiking graph layer and three spatial spiking fully-connected layers, where TLIF neurons that employ the temporal recurrent neuronal dynamics are the basic building blocks. On the other hand, the temporal spiking graph neural network takes as input tactile temporal graphs, and it has one temporal spiking graph layer and three temporal spiking fully-connected layers, where LLIF neurons that possess the spatial recurrent neuronal dynamics are the basic building blocks. 

Based on Eq.~(\ref{e7}), the membrane potential $u_i(t)$ and output spiking state $o_i(t)$ of TLIF neuron $i$ in the spatial spiking graph layer are decided by:
\begin{equation}	\label{e13}
	\begin{split}
		u_i(t) &= \alpha u_i(t - 1)(1 - o_i(t-1)) + I(t),  \\
		o_i(t) &= \begin{cases}
			1 & \text{if $u_i(t) \geq u_{th}$};\\
			0 & \text{otherwise},
		\end{cases}       
	\end{split}
\end{equation}
where $I(t) = \mathbf{GNN}(G_s(t))$ \textbf{is to capture the spatial structural information}. The membrane potential $u_i(t)$ and output spiking state $o_i(t)$ of TLIF neuron $i$ in spatial spiking fully-connected layers are also decided by Eq.~(\ref{e13}), where  $I(t) = \mathbf{FC}(Pre(t))$ and $Pre(t)$ is the previous layer's output at time step $t$. 

Based on Eq.~(\ref{e10}), the membrane potential $u_i(l)$ and output spiking state $o_i(l)$ of LLIF neuron $i$ in the temporal spiking graph layer are decided by:
\begin{equation}	\label{e14}
	\begin{split}
		u_i(l) &= \beta u_i(l_{prev})(1 - o_i(l_{prev})) + I(l),  \\
		o_i(l) &= \begin{cases}
			1 & \text{if $u_i(l) \geq u_{th}$};\\
			0 & \text{otherwise},
		\end{cases}       	
	\end{split}
\end{equation}
where $I(l) = \mathbf{GNN}(G_t(l))$ \textbf{is to model the temporal dependencies in the data}. The membrane potential $u_i(l)$ and output spiking state $o_i(l)$ of LLIF neuron $i$ in temporal spiking fully-connected layers are also decided by Eq.~(\ref{e14}), where  $I(l) = \mathbf{FC}(Pre(l))$ and $Pre(l)$ is the previous layer's output at location $l$. $l$ is the taxel $n\in N$ in event-driven tactile learning. To fairly compare with other baselines, we use TAGConv~\cite{du2017topology} as $\mathbf{GNN}$ in this paper.

The spatial spiking graph neural network finally outputs the spiking feature $O_{1}\in\mathbb{R}^{K\times T}$ and predicts the label vector $O_1'\in\mathbb{R}^{K}$ by averaging $O_1$ over the time window $T$,
\begin{equation}
	\label{e15}
	O_1' = \frac{1}{T}\sum_t^TO_1(t),
\end{equation}
where $O_1(t)\in\mathbb{R}^{K}$. The temporal spiking graph neural network finally outputs the spiking features $O_{2}\in\mathbb{R}^{K\times N}$ and predicts the label vector $O_2'\in\mathbb{R}^{K}$ by averaging $O_2$ over the spatial domain $N$,
\begin{equation}
	\label{e16}
	O_2' = \frac{1}{N}\sum_l^NO_2(l),
\end{equation}
where $O_2(l)\in\mathbb{R}^{K}$.
To fuse the predictions from these two networks, we take the $mean$ or element-wise $max$ of these two label vectors $O_1'$ and $O_2'$ and obtain the final predicted label vector $O'\in\mathbb{R}^{K}$. The predicted label is associated with the neuron with the largest value.


\begin{figure}
     \centering
     \begin{subfigure}[b]{0.24\textwidth}
         \centering
         \includegraphics[width=\textwidth]{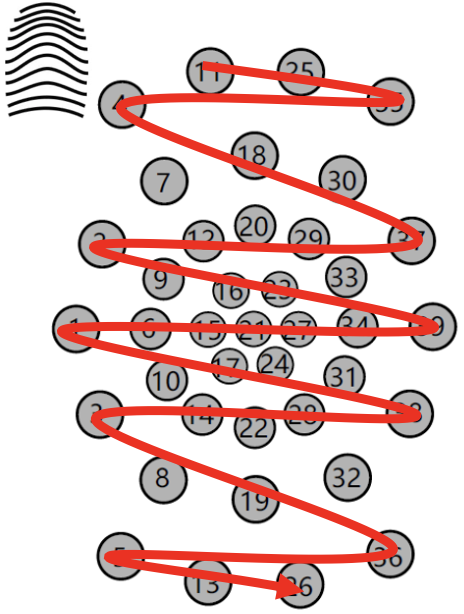}
         \caption{}
         \label{fig:location1}
     \end{subfigure}
     \hfill
     \begin{subfigure}[b]{0.24\textwidth}
         \centering
         \includegraphics[height=47mm, width=\textwidth]{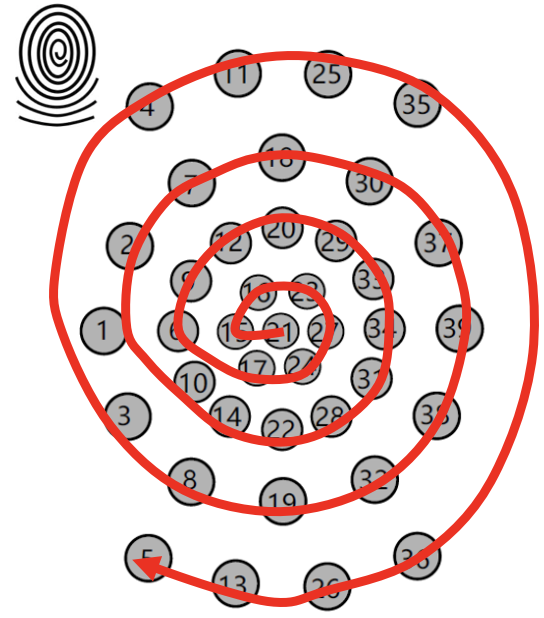}
         \caption{}
         \label{fig:location2}
     \end{subfigure}
     \hfill
     \begin{subfigure}[b]{0.24\textwidth}
         \centering
         \includegraphics[width=\textwidth]{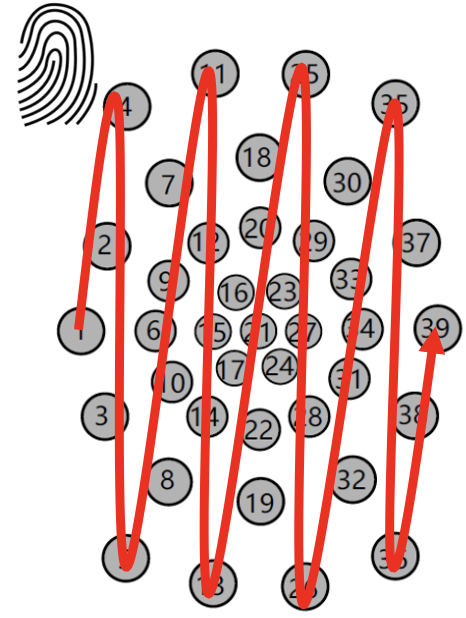}
         \caption{}
         \label{fig:location3}
     \end{subfigure}
     \hfill
     \begin{subfigure}[b]{0.24\textwidth}
         \centering
         \includegraphics[width=\textwidth]{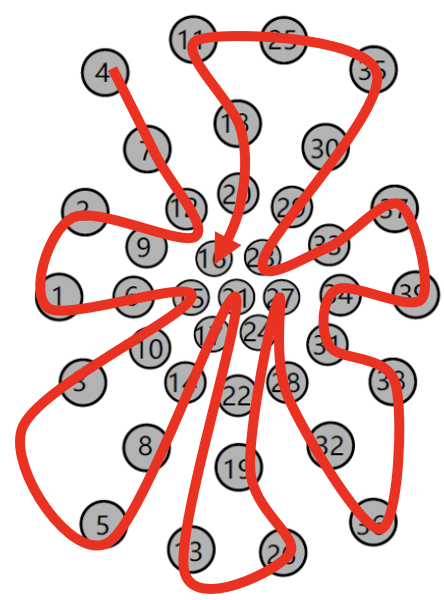}
         \caption{}
         \label{fig:location4}
     \end{subfigure}
        \caption{Location orders. \textbf{(a)} Arch-like location order. \textbf{(b)} Whorl-like location order. \textbf{(c)} Loop-like location order. \textbf{(d)} Random location order.}
        \label{orders}
\end{figure}

	
		

\subsection{Implementations}\label{alg}
In this section, we first introduce the location orders to enable the spatial recurrent neuronal dynamics of location spiking neurons. Then, we present the implementation details and timestep-wise inference algorithms for the proposed models. 
\subsubsection{Location Orders}
\textbf{To enable the spatial recurrent neuronal dynamics of location spiking neurons, we need to manually set the location orders of location spiking neurons.} Specifically, we propose four kinds of location orders for event-driven tactile learning and explore their robustness on the event-driven tactile tasks. As shown in Fig.~\ref{orders}, three location orders are designed based on the major fingerprint patterns of humans -- arch, whorl, and loop. And one location order randomly traverses all the taxels. Four concrete examples are shown below. Each number in the brackets represents the taxel index.
\begin{itemize}
	\item An example for the arch-like location order: [11, 25, 35, 4, 18, 30, 7, 2, 20, 37, 29, 12, 9, 33, 23, 16, 1, 6, 15, 21, 27, 34, 39, 24, 17, 10, 31, 38, 28, 14, 3, 22, 32, 8, 19, 36, 5, 13, 26] 
	\item An example for the whorl-like location order: [21, 15, 16, 23, 27, 24, 17, 6, 9, 12, 20, 29, 33, 34, 31, 28, 22, 14, 10, 1, 2, 7, 18, 30, 37, 39, 38, 32, 19, 8, 3, 4, 11, 25, 35, 36, 26, 13, 5]
	\item An example for the loop-like location order: [1, 2, 3, 4, 5, 6, 7, 8, 9, 10, 11, 12, 13, 14, 15, 16, 17, 18, 19, 20, 21, 22, 23, 24, 25, 26, 27, 28, 29, 30, 31, 32, 33, 34, 35, 36, 37, 38, 39] 
	\item An example for the random location order: [4, 7, 12, 9, 2, 1, 6, 15, 10, 3, 5, 8, 14, 17, 21, 22, 13, 26, 19, 24, 27, 28, 32, 36, 38, 31, 34, 39, 37, 33, 23, 29, 30, 35, 25, 11, 18, 20, 16]
\end{itemize}

\subsubsection{Hybrid\_SRM\_FC}
Similar to the spike-count loss of prior works~\cite{Shrestha2018, taunyazov2020event}, we propose a location spike-count loss to optimize the SNN with LSRM neurons:
\begin{equation}
	\label{e17}
	\mathcal{L}_{LSRM}  =\frac{1}{2}\sum_{k=0}^{K}\left(\sum_{l=0}^{N}o_{k}(l) - \sum_{l=0}^{N}\hat{o}_{k}(l)\right)^2, 
\end{equation}  
which captures the difference between the observed output spike count $\sum_{l=0}^{N}o_{k}(l)$ and the desired spike count $\sum_{l=0}^{N}\hat{o}_{k}(l)$ across the $K$ neurons. Moreover, to optimize the Hybrid\_SRM\_FC, we develop a weighted spike-count loss:
\begin{equation}
	\label{e18}
	\mathcal{L}_{1} = \frac{1}{2}\sum_{k=0}^{K}\left(\left(\sum_{t=0}^{T}o_{k}(t) + \lambda\sum_{l=0}^{N}o_{k}(l)\right) - \sum_{c=0}^{T+ N}\hat{o}_{k}(c)\right)^2, 
\end{equation}  
which first balances the contributions from two SNNs and then captures the difference between the observed balanced output spike count $\sum_{t=0}^{T}o_{k}(t) + \lambda\sum_{l=0}^{N}o_{k}(l)$ and the desired spike count $\sum_{c=0}^{T+ N}\hat{o}_{k}(c)$ across the $K$ output neurons. For both $\mathcal{L}_{LSRM}$ and $\mathcal{L}_{1}$, the desired spike counts have to be specified for the correct and incorrect classes and are task-dependent hyperparameters. We set these hyperparameters as in~\cite{taunyazov2020event}. To overcome the non-differentiability of spikes and apply the backpropagation algorithm, we use the approximate gradient proposed in SLAYER~\cite{Shrestha2018}. Moreover, based on the SLAYER's weight update in the temporal domain, we can derive the weight update for the SNNs with LSRM neurons in the spatial domain. Please check more details in our Github repository.

To demonstrate the applicability of our model to the spike-based temporal data, we propose the timestep-wise inference algorithm of the Hybrid\_SRM\_FC, which is shown in Algorithm~\ref{alg:timestep}. The corresponding timestep-wise training algorithm can be derived by incorporating the weighted spike-count loss.
\begin{algorithm}[htp]
	\caption{Timestep-wise inference algorithm of the Hybrid\_SRM\_FC, adopted from~\cite{kangTactile}}
	\label{alg:timestep}
	\begin{algorithmic}[1]
		\Require{event-based tactile inputs $X_{in} \in \mathbb{R}^{N\times T}$, $N$ taxels, and the total time length $T$.} 
		\Ensure{timestep-wise predictions of $O_{1}$, $O_{2}$, and $O$.}
		\For{$t \gets 1$ to $T$}     
		\State {obtain $X\in \mathbb{R}^{N\times t}$}  
		\State {obtain $\bar{X'}=concatenate(X', \mathbf{0}) \in\mathbb{R}^{T\times N}$, where $X'\in\mathbb{R}^{t\times N}$, and $\mathbf{0}\in\mathbb{R}^{(T-t)\times N}$}  
		\State {$O_{1}(t)=\mathbf{0}\in\mathbb{R}^{K\times t}$, $O_{2}(t)=\mathbf{0}\in\mathbb{R}^{K\times N}$}  
		\State {$O(t)=\mathbf{0}\in\mathbb{R}^{K\times (t+N)}$}  
		\State {$O_{1}(t)$ = SNN\_TSRM($X$)} \Comment{SNN\_TSRM for the fully-connected SNN with TSRM neurons}
		\State {$O_{2}(t)$ = SNN\_LSRM($\bar{X'}$)} \Comment{SNN\_LSRM for the fully-connected SNN with LSRM neurons}
		\State {$O(t)$ = $concatenate$($O_{1}(t), O_{2}(t)$) }
		\EndFor              
	\end{algorithmic}
\end{algorithm}

\subsubsection{Hybrid\_LIF\_GNN}
To train the Hybrid\_LIF\_GNN, we define the loss function that captures the mean squared error between the ground truth label vector $y$ and the final predicted label vector $O'$. 
\begin{equation}
	\label{e19}
	\mathcal{L}_{2}  = \|y - O'\|^2.
\end{equation}  
We utilize the spatio-temporal backpropagation~\cite{wu2018spatio} to derive the weight update for the SNNs with LLIF neurons. Moreover, to overcome the non-differentiability of spikes, we use the rectangular function~\cite{wu2018spatio} to approximate the derivative of the spike function (Heaviside function) in Eqs.~(\ref{e13}) and (\ref{e14}). Please check more implementation details in our Github repository. Algorithm~\ref{alg:timestep_spikingfusion} presents the timestep-wise inference algorithm of the Hybrid\_LIF\_GNN.
\begin{algorithm}
	\caption{Timestep-wise inference algorithm of the Hybrid\_LIF\_GNN}
	\label{alg:timestep_spikingfusion}
	\begin{algorithmic}[1]
		\Require{event-based tactile inputs $X_{in} \in \mathbb{R}^{N\times T}$, $N$ taxels, and the total time length $T$} 
		\Ensure{timestep-wise label vectors of $O_{1}'$, $O_{2}'$, and $O'$}
		\For{$t \gets 1$ to $T$}     
		\State {form $t$ tactile spatial graphs $G_s$ with $X\in \mathbb{R}^{N\times t}$}  
		\State {obtain $\bar{X'}=concatenate(X', \mathbf{0}) \in\mathbb{R}^{T\times N}$, where $X'\in\mathbb{R}^{t\times N}$, and $\mathbf{0}\in\mathbb{R}^{(T-t)\times N}$}  
		\State {form $N$ tactile temporal graphs $G_t$ with $\bar{X'}$}  
		\State {$O_1'(t), O_2'(t), O'(t)=\mathbf{0}\in\mathbb{R}^K$}  
		\For{$i \gets 1$ to $t$}    
		\State {$O_1'(t)$ += SSGNN($G_s(i)$) } \Comment{SSGNN for the spatial spiking graph neural network}
		\EndFor
		\State {$O_1'(t)$ /= $t$ }
		\For{$j \gets 1$ to $N$}    
		\State {$O_2'(t)$ += TSGNN($G_t(j)$) }  \Comment{TSGNN for the temporal spiking graph neural network}
		\EndFor
		\State {$O_2'(t)$ /= $N$ }
		\State {$O'(t)$ = $mean$($O_1'(t), O_2'(t)$) }\Comment{$max$ can be used}
		\EndFor              
	\end{algorithmic}
\end{algorithm}

\section{Experiments}\label{sec:exp}

We extensively evaluate our proposed models and demonstrate their effectiveness and efficiency on event-driven tactile learning, including event-driven tactile object recognition and event-driven slip detection. Specifically, we first conduct experiments on the Hybrid\_SRM\_FC to show that location spiking neurons can improve event-driven tactile learning. Then, we utilize the experiments on the Hybrid\_LIF\_GNN to show that location spiking neurons are user-friendly and can be incorporated into more powerful spike-based learning frameworks to further boost event-driven tactile learning. The source code and experimental configuration details are available at \url{https://github.com/pkang2017/TactileLSN}.

\subsection{Hybrid\_SRM\_FC}
In this section, we first introduce the datasets and models for the experiments. Next, to show the effectiveness of the Hybrid\_SRM\_FC, we extensively evaluate it on the benchmark datasets and compare it with state-of-the-art models. Finally, we demonstrate the superior energy efficiency of the Hybrid\_SRM\_FC over the counterpart ANNs and show the high-efficiency benefit of LSRM neurons. We implement our models using slayerPytorch\footnote{https://github.com/bamsumit/slayerPytorch} and employ RMSProp with the $l_2$ regularization to optimize them. 


\subsubsection{Datasets}
We use the datasets collected by NeuTouch~\cite{taunyazov2020event}, including ``Objects-v1'' and ``Containers-v1'' for event-driven tactile object recognition and ``Slip Detection'' for event-driven slip detection. Unlike ``Objects-v1'' which only requires models to determine the type of objects being handled, ``Containers-v1'' asks models about the type of containers being handled and the amount of liquid (0\%, 25\%, 50\%, 75\%, 100\%) held within. Thus, ``Containers-v1'' is more challenging for event-driven tactile object recognition. Moreover, the task of event-driven slip detection is also challenging since it requires models to detect the rotational slip within a short time, like 0.15s for ``Slip Detection''. We provide more details about the datasets in the Supplementary Material. Following the experimental setting of~\cite{taunyazov2020event}, we split the data into a training set (80\%) and a testing set (20\%), repeat each experiment for five rounds, and report the average accuracy.



\subsubsection{Comparing Models}
We compare our model with the state-of-the-art SNN methods for event-driven tactile learning, including Tactile-SNN~\cite{taunyazov2020event} and TactileSGNet~\cite{gu2020tactilesgnet}. Tactile-SNN employs \textbf{TSRM neurons} as the building blocks, and the network structure of Tactile-SNN is the same as the fully-connected SNN with TSRM neurons in the Hybrid\_SRM\_FC. TactileSGNet utilizes \textbf{TLIF neurons} as the building blocks and the network structure of TactileSGNet is the same as the spatial spiking graph neural network in the Hybrid\_LIF\_GNN. As in~\cite{taunyazov2020event}, we also compare our model against conventional deep learning, specifically Gated Recurrent Units (GRUs)~\cite{cho2014learning} with Multi-layer Perceptrons (MLPs) and 3D convolutional neural networks (3D\_CNN)~\cite{gandarias2019active}. The network structure of GRU-MLP is Input-GRU-MLP, where MLP is only utilized at the final time step. And the network structure of CNN-3D is Input-3D\_CNN1-3D\_CNN2-FC, where FC is for the fully-connected layer.

\begin{table}[]
\captionsetup{singlelinecheck = false, justification=justified}
	\begin{threeparttable}
		\caption{Accuracies on benchmark datasets for the Hybrid\_SRM\_FC}
		\setlength{\tabcolsep}{3.2mm}{
			\begin{tabular}{lcccc}
				\hline
				Method       & Type & Objects-v1       & Containers-v1    & Slip Detection \\ \hline
				Tactile-SNN~\cite{taunyazov2020event}       & SNN  & 0.75          & \hspace{1.5mm}0.57*          & \hspace{1.5mm}0.82*           \\ \hline
				TactileSGNet~\cite{gu2020tactilesgnet} & SNN  & 0.79          & 0.58          & 0.97           \\ \hline
				GRU-MLP~\cite{taunyazov2020event}      & ANN  & 0.72          & \hspace{1.5mm}0.46*          & \hspace{1.5mm}0.87*           \\ \hline
				CNN-3D~\cite{taunyazov2020event}       & ANN  & 0.90          & \hspace{1.5mm}0.67*          & \hspace{1.5mm}0.44*           \\ \hline
				Hybrid\_SRM\_FC  & SNN  & \textbf{0.91} & \textbf{0.86} & \textbf{1.0}   \\ \hline
		\end{tabular}}\label{t2}
		\begin{tablenotes}
			\item *These values come from~\cite{taunyazov2020event}. The best performance is in bold.
		\end{tablenotes}
	\end{threeparttable}
\end{table}

\subsubsection{Basic Performance}
Table~\ref{t2} presents the test accuracies on the three datasets. We observe that the Hybrid\_SRM\_FC significantly outperforms the state-of-the-art SNNs. The reason why our model is superior to other SNNs could be two-fold: (1) different from state-of-the-art SNNs that only extract features with existing spiking neurons, our model employs an SNN with location spiking neurons that enhance the representative ability and enable the model to extract features in a novel way; (2) our model fuses the SNN with TSRM neurons and the SNN with LSRM neurons to better capture complex spatio-temporal dependencies in the data. We also compare our model with ANNs, which provide fair comparison baselines for fully ANN architectures since they employ similar lightsome network architectures as ours. From Table~\ref{t2}, we find out that our model outperforms the counterpart ANNs on the three tasks, which might be because our model is more compatible with event-based tactile data and better maintains the sparsity to prevent overfitting.

\subsubsection{Ablation Studies}
To examine the effectiveness of each component in the proposed model and validate the representation ability of location spiking neurons on event-driven tactile learning, we \textbf{separately train} the SNN with TSRM neurons (which is exactly Tactile-SNN) and the SNN with LSRM neurons (which is referred to as Location Tactile-SNN). From Table~\ref{t3}, we surprisingly find out that Location Tactile-SNN significantly surpasses Tactile-SNN on the datasets for event-driven tactile object recognition and provides comparable performance on event-driven slip detection. The reason for this could be two-fold: (1) the time durations of event-driven tactile object recognition datasets are longer than that of ``Slip Detection'', and Location Tactile-SNN with LSRM neurons is good at capturing the mid-and-long term dependencies in these object recognition datasets; (2) like Tactile-SNN, Location Tactile-SNN with LSRM neurons can still capture the spatial dependencies in the event-driven tactile data (``Slip Detection'') due to the spatial recurrent neuronal dynamics of location spiking neurons. Moreover, we examine the sensitivity of $\lambda$ in Eq.(\ref{e18}) and the robustness of location orders. From Table~\ref{t3}, we notice the results of related models are close, proving that the $\lambda$ tuning and location orders do not significantly impact task performance.

\begin{table}[]
\captionsetup{singlelinecheck = false, justification=justified}
	\caption{Ablation studies on the Hybrid\_SRM\_FC}
	\setlength{\tabcolsep}{3.2mm}{
		\begin{tabular}{lcccc}
			\hline
			Method               & Type & Objects-v1 & Containers-v1 & Slip Detection \\ \hline
			Tactile-SNN~\cite{taunyazov2020event}               & SNN  & 0.75    & 0.57       & 0.82           \\ \hline
			Location Tactile-SNN & SNN  & 0.89    & 0.88       & 0.82           \\ \hline
			Hybrid\_SRM\_FC $\lambda=1$          & SNN  & 0.91    & 0.86       & 1.0            \\ \hline
			Hybrid\_SRM\_FC $\lambda=0.5$         & SNN  & 0.92    & 0.89       & 0.98           \\ \hline
			Hybrid\_SRM\_FC-loop  & SNN  & 0.91    & 0.86       & 1.0            \\ \hline
			Hybrid\_SRM\_FC-arch  & SNN  & 0.91    & 0.86       & 0.99           \\ \hline
			Hybrid\_SRM\_FC-whorl & SNN  & 0.92    & 0.86       & 0.98           \\ \hline
			Hybrid\_SRM\_FC-random & SNN  & 0.91    & 0.86       & 0.99           \\ \hline
	\end{tabular}}\label{t3}
\end{table}

\subsubsection{Timestep-wise Inference}
\begin{figure}
     \centering
     \begin{subfigure}[b]{0.3\textwidth}
         \centering
         \includegraphics[width=\textwidth]{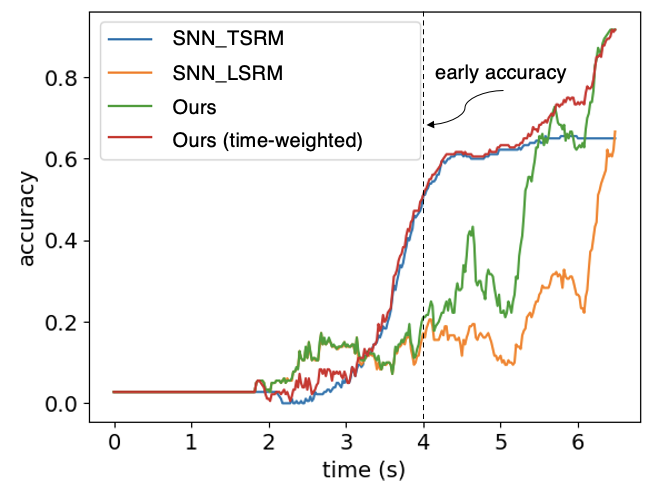}
         \caption{}
         \label{fig:ts1}
     \end{subfigure}
     \hfill
     \begin{subfigure}[b]{0.3\textwidth}
         \centering
         \includegraphics[width=\textwidth]{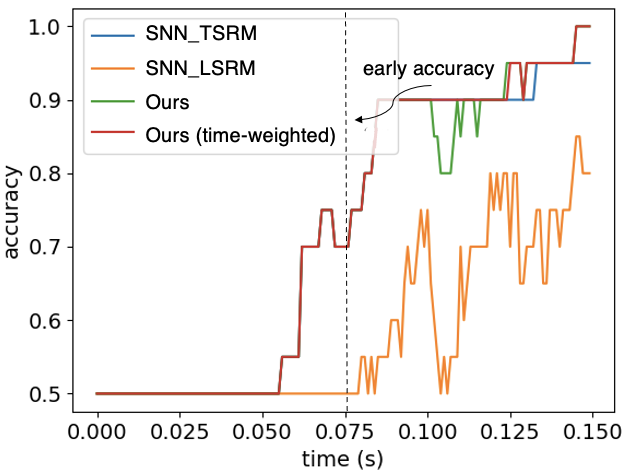}
         \caption{}
         \label{fig:ts2}
     \end{subfigure}
     \hfill
     \begin{subfigure}[b]{0.3\textwidth}
         \centering
         \includegraphics[width=\textwidth]{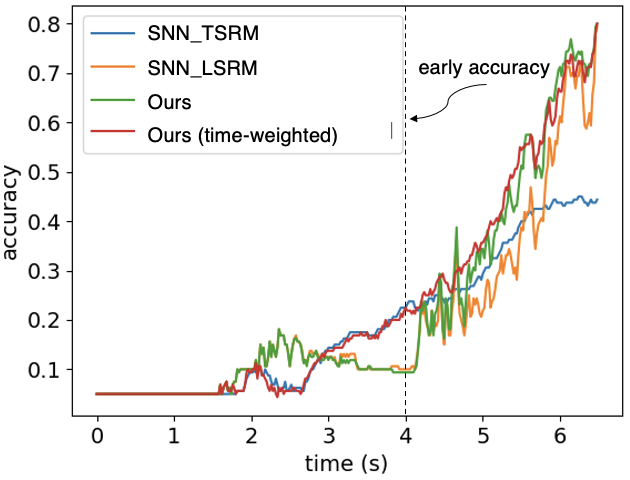}
         \caption{}
         \label{fig:ts3}
     \end{subfigure}
        \caption{The timestep-wise inference (Alg.~\ref{alg:timestep}) for the SNN with TSRM neurons (SNN\_TSRM), the SNN with LSRM neurons (SNN\_LSRM), the Hybrid\_SRM\_FC, and the time-weighted Hybrid\_SRM\_FC on \textbf{(a)} ``Objects-v1'', \textbf{(b)} ``Slip Detection'', \textbf{(c)} ``Containers-v1''. Please note that we use the same event sequences as~\cite{taunyazov2020event} and the first spike occurs at around 2.0s for ``Objects-v1'' and ``Containers-v1''.}
        \label{fig:ts}
\end{figure}

We evaluate the timestep-wise inference performance of the Hybrid\_SRM\_FC and validate the contributions of the two components in it. Moreover, we propose a time-weighted Hybrid\_SRM\_FC to better balance the two components' contributions and achieve better overall performance. Figure~\ref{fig:ts1}, \ref{fig:ts2}, and \ref{fig:ts3} show the timestep-wise inference accuracies of the SNN with TSRM neurons, the SNN with LSRM neurons, the Hybrid\_SRM\_FC, and the time-weighted Hybrid\_SRM\_FC on the three datasets. Specifically, the output of the time-weighted Hybrid\_SRM\_FC at time $t$ is 
\begin{equation}
	\label{e20}
	\begin{split}
		O_{tw}(t) = conca&tenate((1-\omega) * O_1(t), \omega * O_2(t)), \\
		\omega &= \frac{1}{1+e^{-\psi*(\frac{t}{T}-1) }},
	\end{split}
\end{equation}
where the hyperparameter $\psi$ balances the contributions of the two components in the hybrid model and $T$ is the total time length. From the figures, we can see that the SNN with TSRM neurons has good ``early'' accuracies on the three tasks since it well captures the spatial dependencies with the help of Eq.~(\ref{e11}). However, its accuracies do not improve too much at the later stage since it does not sufficiently capture the temporal dependencies. In contrast, the SNN with LSRM neurons has fair ``early'' accuracies, while its accuracies jump a lot at the later stage since it models the temporal dependencies in Eq.~(\ref{e12}). The Hybrid\_SRM\_FC adopts the advantages of these two components and extracts spatio-temporal features from various views, which enables it to have a better overall performance. Furthermore, after employing the time-weighted output and shifting more weights to the SNN with TSRM neurons at the early stage, the time-weighted Hybrid\_SRM\_FC can have a good ``early'' accuracy as well as an excellent ``final'' accuracy.

\subsubsection{Energy Efficiency}\label{e-efficiency-lsrm}

To further analyze the benefits of the proposed model and location spiking neurons, we estimate the gain in computational costs compared to fully ANN architectures. Typically, the number of
synaptic operations is used as a metric for benchmarking the computational energy of SNN models \cite{ijcai2021-441, lee2020spike}. In addition, we can estimate the total energy consumption of a model based on CMOS technology~\cite{horowitz20141}.

Different from ANNs that always conduct real-valued matrix-vector multiplication operations without considering the sparsity of inputs, SNNs carry out event-based computations only
at the arrival of input spikes. Hence, we first measure the mean spiking rate of layer $l$ in our proposed model. Specifically, the mean spiking rate of the layer $l$ in the SNN with existing spiking neurons is given by:
\begin{equation}
	\label{e-fr}
F_{1}^{(l)} = \frac{1}{T}\sum_{t\in T}\frac{\text{\#spikes of layer $l$ at time $t$}}{\text{\#neurons of layer $l$}},
\end{equation}
where $T$ is the total time length. And the mean spiking rate of the layer $l$ in the SNN with location spiking neurons is given by:
\begin{equation}
	\label{e-fr}
	F_{2}^{(l)} = \frac{1}{N}\sum_{n\in N}\frac{\text{\#spikes of layer $l$ at location $n$}}{\text{\#neurons of layer $l$}},
\end{equation}
where $N$ is the total number of locations. We show the mean spiking rates of Hybrid\_SRM\_FC layers in the Supplementary Material. With the mean spiking rates, we can estimate the number of synaptic operations in the SNNs. Given $M$ is the number of neurons, $C$ is the number of synaptic connections per neuron, and $F$ indicates the mean spiking rate, the number of synaptic operations at each time or location in layer $l$ is calculated as $M^{(l)}\times C^{(l)}\times F^{(l)}$, where $F^{(l)}$ is $F_1^{(l)}$ or $F_2^{(l)}$. Thus, the total number of synaptic operations in our hybrid model is calculated by: 
\begin{equation}
	\label{e-fr}
	OP_{Hybrid} = \sum_{l}M^{(l)}\times C^{(l)}\times F_{1}^{(l)}\times T + \sum_{l'}M^{(l')}\times C^{(l')}\times F_{2}^{(l')}\times N,
\end{equation}
where $l$ is the spiking layer with existing spiking neurons and $l'$ is the spiking layer with location spiking neurons. Generally, the total number of synaptic operations in the ANNs is $\sum_lM^{(l)}\times C^{(l)}$. Based on these, we estimate the number of synaptic operations in the Hybrid\_SRM\_FC and ANNs like the GRU-MLP and CNN-3D. As shown in Table~\ref{t4}, all the SNNs achieve far fewer operations than ANNs on the three datasets. 

Moreover, due to the binary nature of spikes, SNNs perform only accumulation (AC) per synaptic operation, while ANNs perform the multiply-accumulate (MAC) computations since the operations are real-valued. In general, AC computation is considered to be significantly more energy-efficient than MAC. For example, an AC is reported to be $\mathbf{5.1\times}$ more energy-efficient than a MAC in the case of 32-bit floating-point numbers (45nm CMOS process)~\cite{horowitz20141}. Based on this principle, we obtain the computational energy benefits of SNNs over ANNs in Table~\ref{t4}. From the table, we can see that the SNN models are \textbf{10$\times$} to \textbf{100$\times$} more energy-efficient than ANNs and the location spiking neurons (LSRM neurons) have the similar energy efficiency compared to existing spiking neurons (TSRM neurons). 

These results are consistent with the fact that the sparse spike communication and event-driven computation underlie the efficiency advantage of SNNs and demonstrate the potential of our model and location spiking neurons on neuromorphic hardware.



\begin{table}[ht]
	\caption{The number of synaptic operations ($\#op$, $\times 10^6$) and the compute-energy benefit (the compute-energy of ANNs / the compute-energy of SNNs, 45nm) on benchmark datasets for the Hybrid\_SRM\_FC}
	\setlength{\tabcolsep}{2.6mm}{
		\begin{tabular}{lcccc}
			\hline
			Method   & Type   & Objects-v1                               & Containers-v1                             & Slip Detection                          \\ \hline
			$\#op$ GRU-MLP  & ANN   & 5.89                                 & 5.89                                 & 2.72                                   \\ \hline
			$\#op$ CNN-3D    &ANN  & 4.17                                  & 4.07                                  & 1.75                                     \\ \hline\hline
			$\#op$ SNN with TSRM neurons &SNN& 0.31                                  & 0.42                                   & 0.022                                   \\ 
			Compute-energy Benefit & & 68.60$\sim$96.90$\times$ & 49.42$\sim$71.52$\times$ & 405.68$\sim$630.55$\times$ \\ \hline
			$\#op$ SNN with LSRM neurons &SNN& 0.29                                  & 0.41                                   & 0.023                                   \\ 
			Compute-energy Benefit & & 73.33$\sim$103.58$\times$ & 50.63$\sim$73.27$\times$ & 388.04$\sim$603.13$\times$ \\ \hline
			$\#op$ Hybrid\_SRM\_FC  &SNN& 0.60                                  & 0.83                                   & 0.045                                   \\ 
			Compute-energy Benefit & &35.45$\sim$50.07$\times$ & 25.01$\sim$36.19$\times$ & 198.33$\sim$308.27$\times$ \\ \hline
	\end{tabular}}\label{t4}
\end{table}

\subsection{Hybrid\_LIF\_GNN}

In this section, to show the usability of location spiking neurons and further boost event-driven tactile learning, we conduct a series of experiments with the Hybrid\_LIF\_GNN, which is powered by the popular spike-based learning framework -- STBP~\cite{wu2018spatio}. Specifically, we first compare our model with the state-of-the-art models with TLIF neurons and GNN structures. Then, we conduct several ablation studies to examine the effectiveness of some designs in the Hybrid\_LIF\_GNN. Next, we demonstrate the superior energy efficiency of our model over the counterpart Graph Neural Networks (GNNs) and show the high-efficiency benefits of location spiking neurons. Finally, we compare with the Hybrid\_SRM\_FC on the same benchmark datasets to validate the superiority of the Hybrid\_LIF\_GNN.\footnote{In this section, to be consistent with~\cite{gu2020tactilesgnet}, we use accuracies (\%).} 

\subsubsection{Datasets}

To fairly compare with other published models with TLIF neurons~\cite{gu2020tactilesgnet}, we evaluate the Hybrid\_LIF\_GNN on ``Objects-v0'' and ``Containers-v0''. These two datasets are the initial versions of ``Objects-v1'' and ``Containers-v1''. We  demonstrate their differences in the Supplementary Material. To show the superiority of the Hybrid\_LIF\_GNN on event-driven tactile learning, we compare it with the Hybrid\_SRM\_FC on ``Objects-v1'', ``Containers-v1'', and ``Slip Detection''. During the experiments, we split the data into a training set (80\%) and a testing set (20\%) with an equal class distribution. We repeat each experiment for five rounds and report the average accuracy. 



\subsubsection{Comparing Models}

We compare the Hybrid\_LIF\_GNN with the state-of-the-art methods with TLIF neurons and GNN structures~\cite{gu2020tactilesgnet} on event-based tactile object recognition. Specifically, we compare the TactileSGNet series. The general network structure is the same as the spatial spiking graph neural network, which is Input-Spiking TAGConv-Spiking FC1-Spiking FC2-Spiking FC3. The other models in the series are obtained by substituting the Spiking TAGConv layer:
\begin{itemize}
	\item TactileSGNet-MLP, which uses the Spiking FC layer with TLIF neurons to process the input. The network structure is Input-Spiking FC0-Spiking FC1-Spiking FC2-Spiking FC3.
	\item TactileSGNet-CNN, which takes the network structure of Input-Spiking CNN-Spiking FC1-Spiking FC2-Spiking FC3. The tactile input is organized in a grid structure according to the spatial distribution of taxels, and the Spiking CNN with TLIF neurons is utilized to extract features from this grid. 
	\item TactileSGNet-GCN, where the graph convolutional network (GCN) is used as the GNN in Eq.~(\ref{e13}). The network structure is Input-Spiking GCN-Spiking FC1-Spiking FC2-Spiking FC3.
\end{itemize}
Moreover, we also compare the Hybrid\_LIF\_GNN against fully GNNs. Specifically, \textbf{the GNNs have the same network structures as the Hybrid\_LIF\_GNN}, including one recurrent TAGConv-FC1-FC2-FC3 for $T$ tactile spatial graphs, one recurrent TAGConv-FC1-FC2-FC3 for $N$ tactile temporal graphs, and one fusion module to fuse the predictions from two branches. The major difference between our model and GNNs is that GNNs employ artificial neurons and adopt different activation functions in Eqs.~(\ref{e13}) and (\ref{e14}) while our model utilizes the spiking neurons and takes the Heaviside function as the activation function.

\begin{table}[]
\captionsetup{singlelinecheck = false, justification=justified}
	\begin{threeparttable}
		\caption{Accuracies (\%) on datasets for the Hybrid\_LIF\_GNN}
		\label{tab1}
		\setlength{\tabcolsep}{5.5mm}{
		\begin{tabular}{lccc}
			\hline
			Method                & \multicolumn{1}{l}{Type} & \multicolumn{1}{l}{Objects-v0} & \multicolumn{1}{l}{Containers-v0}  \\\hline
			TactileSGNet-MLP~\cite{gu2020tactilesgnet}      & SNN                      & \hspace{2mm}85.97*                       & \hspace{2mm}58.83*                           \\\hline
			TactileSGNet-CNN~\cite{gu2020tactilesgnet} & SNN                      & \hspace{2mm}88.40*                       & \hspace{2mm}60.17*                           \\\hline
			TactileSGNet-GCN~\cite{gu2020tactilesgnet}      & SNN                      & \hspace{2mm}85.14*                       & \hspace{2mm}58.83*                           \\\hline
			TactileSGNet-TAGConv~\cite{gu2020tactilesgnet}          & SNN                      & \hspace{2mm}89.44*                       & \hspace{2mm}64.17*                           \\\hline
			Recurrent GNN-linear            & GNN                      & 92.36                       & 70.67                        \\\hline
			Recurrent GNN-elu               & GNN                      & 91.11                       & 74.67                         \\\hline
			Recurrent GNN-LeakyRelu         & GNN                      & 89.31                       & 73.00                       \\\hline \hline
			Hybrid\_LIF\_GNN-sparse-mean          & SNN                      & \textbf{93.33}              & \textbf{79.33}         \\\hline
			Hybrid\_LIF\_GNN-dense-mean          & SNN               & 92.50                      & 78.67                           \\\hline
			Hybrid\_LIF\_GNN-sparse-max         & SNN          & 85.56                      & 77.00                        \\\hline
			Hybrid\_LIF\_GNN-dense-max        & SNN           & 85.14                     & 76.00                       \\ \hline
		\end{tabular}}
		\begin{tablenotes}
			\item *These values come from~\cite{gu2020tactilesgnet}. All the Hybrid\_LIF\_GNN models use the loop-like location order.  ``sparse'' is for ``sparse tactile temporal graph'', ``dense'' is for ``dense tactile temporal graph'', ``mean'' is for ``mean fusion'', and ``max'' is for ``max fusion''. The best performance is in bold. 
		\end{tablenotes}
	\end{threeparttable}
\end{table}
\subsubsection{Basic Performance} We report the test accuracies on the two event-driven tactile object recognition datasets in Table~\ref{tab1}. From this table, we can see that the Hybrid\_LIF\_GNN significantly outperforms the TactileSGNet series~\cite{gu2020tactilesgnet}. The reason why our model can achieve the better performance could be two-fold: (1) different from the TactileSGNet models that only utilize TLIF neurons to extract features from the tactile spatial graphs, our model also employs the temporal spiking graph neural network with LLIF neurons to extract features from the tactile temporal graphs; (2) our model fuses the spatial and temporal spiking graph neural networks to capture complex spatio-temporal dependencies in the data. We also compare our model with fully GNNs by replacing the spike functions in Eqs.~(\ref{e13}) and (\ref{e14}) with activation functions, such as linear, elu, or LeakyRelu. These models provide fair comparison baselines for fully GNN architectures since they employ the same network architecture as ours. From Table~\ref{tab1}, we observe that the Hybrid\_LIF\_GNN outperforms the counterpart GNNs on the two datasets, which might be because our model is more compatible with event-based tactile data and better maintains the sparsity to prevent overfitting. 
\subsubsection{Ablation Studies} We further provide ablation studies for exploring the optimal design choices. From Table~\ref{tab1}, we find out that the combination of ``sparse tactile temporal graph'' and ``mean fusion'' performs better than other combinations. The reason for this could be two-fold: (1) the dense tactile temporal graph involves too many insignificant temporal dependencies and does not differentiate the importance of each dependency; (2) the max fusion results in information loss. 

%

\subsubsection{Timestep-wise Inference} 
\begin{figure}
     \centering
     \begin{subfigure}[b]{0.3\textwidth}
         \centering
         \includegraphics[width=\textwidth]{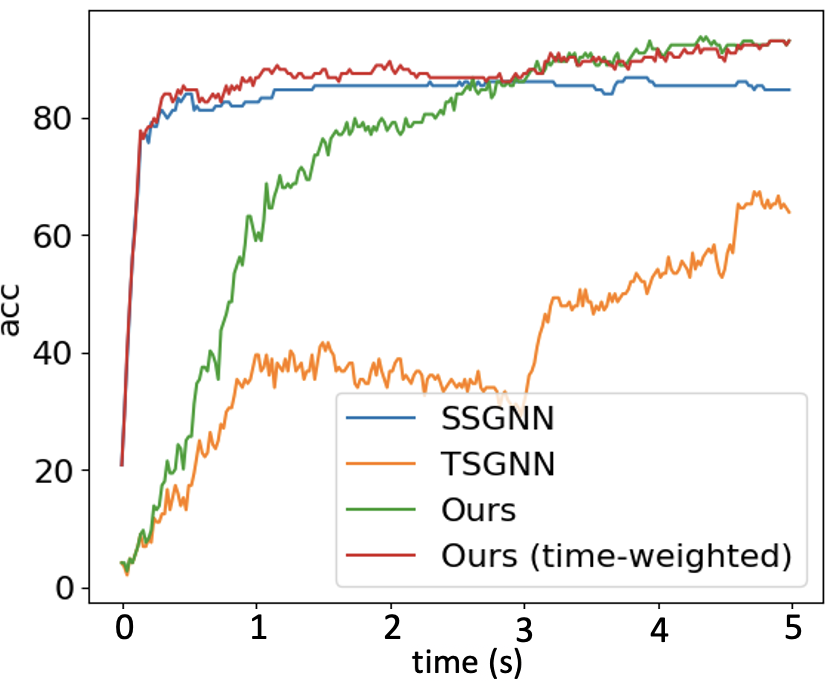}
         \caption{}
         \label{fig:ts4}
     \end{subfigure}
     \hspace{5mm}
     \begin{subfigure}[b]{0.3\textwidth}
         \centering
         \includegraphics[width=\textwidth]{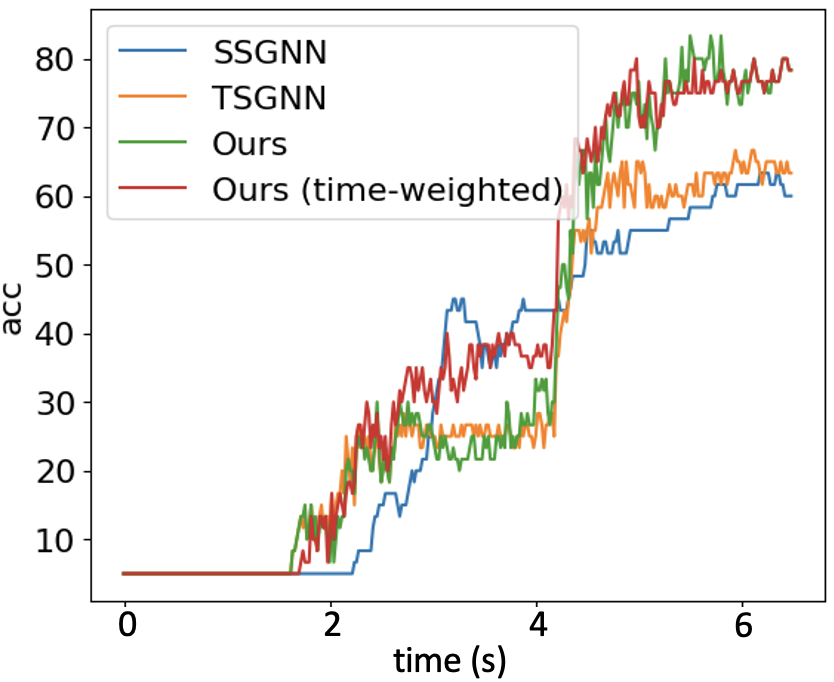}
         \caption{}
         \label{fig:ts5}
     \end{subfigure}
        \caption{The timestep-wise inference (Alg.~\ref{alg:timestep_spikingfusion}) accuracies (\%) for the spatial spiking graph neural network (SSGNN), the temporal spiking graph neural network (TSGNN), the Hybrid\_LIF\_GNN, and the time-weighted Hybrid\_LIF\_GNN on \textbf{(a)} ``Objects-v0'', \textbf{(b)} ``Containers-v0''.}
        \label{fig:ts-gnn}
\end{figure}

Figure~\ref{fig:ts-gnn} shows the timestep-wise inference accuracies (\%) for the spatial spiking graph neural network, the temporal spiking graph neural network, the Hybrid\_LIF\_GNN, and the time-weighted Hybrid\_LIF\_GNN on the two datasets. Specifically, the output of time-weighted Hybrid\_LIF\_GNN at time $t$ is 
\begin{equation}
	\label{e21}
	O'_{tw}(t) = O_1'(t)(1-\frac{t}{\zeta T}) + O_2'(t)\frac{t}{\zeta T},
\end{equation}
where $\zeta$ balances the contributions of the two components in the hybrid model and $T$ is the total time length. From the figure, we can see that the spatial spiking graph neural network has a good ``early'' accuracy with the help of tactile spatial graphs, while its accuracy does not improve too much at the later stage since it cannot well capture the temporal dependencies. In contrast, the temporal spiking graph neural network has a fair ``early'' accuracy, while its accuracy jumps a lot at the later stage since it models the temporal dependencies explicitly. The Hybrid\_LIF\_GNN adopts the advantages of these two models and extracts spatio-temporal features from multiple views, which enables it to have a better overall performance. Furthermore, after employing the time-weighted output and setting $\zeta=2$ to shift more weights to the spatial spiking graph neural network at the early stage, the time-weighted model can have a good ``early'' accuracy as well as an excellent ``final'' accuracy, see red lines in Fig. \ref{fig:ts-gnn}.

\subsubsection{Energy Efficiency}
Following the estimation methods in Section~\ref{e-efficiency-lsrm}, we estimate the computational costs of the Hybrid\_LIF\_GNN and its counterpart GNNs on the benchmark datasets. 

We show the mean spiking rates of Hybrid\_LIF\_GNN layers in the Supplementary Material. Table~\ref{t-e} provides the number of synaptic operations conducted in the Hybrid\_LIF\_GNN and the counterpart GNNs with the same network structure. From the table, we can see that the SNNs achieve far fewer operations than GNNs on the benchmark datasets. Moreover, following the 45nm CMOS technology energy principle in Section~\ref{e-efficiency-lsrm}, we obtain the computational energy benefits of SNNs over GNNs in Table~\ref{t-e}. From the table, we can see that the SNN models are \textbf{10$\times$} to \textbf{100$\times$} energy-efficient than GNNs. Furthermore, by comparing the number of synaptic operations in the spatial spiking graph neural network with that in the temporal spiking graph neural network, we find that the temporal spiking graph neural network has the higher energy efficiency. The reason for this could be that we employ the \textbf{sparse tactile temporal graphs} in the temporal spiking graph neural network and such graphs require fewer operations.

These results are consistent with what we show in Section~\ref{e-efficiency-lsrm} and demonstrate the potential of our models and location spiking neurons (LLIF neurons) on neuromorphic hardware.



\begin{table}[ht]
	\caption{The number of synaptic operations ($\#op$, $\times 10^8$) and the compute-energy benefit (the compute-energy of GNNs / the compute-energy of SNNs, 45nm) on benchmark datasets for the Hybrid\_LIF\_GNN}
	\setlength{\tabcolsep}{5mm}{
		\begin{tabular}{lccc}
			\hline
			Method     &Type & Objects-v0                               & Containers-v0                         \\ \hline
			$\#op$ Recurrent GNNs in Table~\ref{tab1}   &GNN     & 1.7188                                 & 2.2146                              \\ \hline\hline
			$\#op$ Spatial spiking graph neural network & SNN& 0.1132                  & 0.1023         \\
			Compute-energy Benefit & & \hspace{1mm}77.44$\times$ & \hspace{3mm}110.41$\times$ \\ \hline
			$\#op$ Temporal spiking graph neural network &SNN & 0.0297           & 0.0313                            \\ 
			Compute-energy Benefit & & \hspace{3mm}295.15$\times$ & \hspace{3mm}360.85$\times$\\ \hline
			$\#op$ Hybrid\_LIF\_GNN & SNN& 0.1429                       & 0.1336            \\ 
			Compute-energy Benefit & & \hspace{1mm}61.34$\times$ & \hspace{1mm}84.54$\times$ \\ \hline
	\end{tabular}}\label{t-e}
\end{table}



\subsubsection{Performance Comparison with the Hybrid\_SRM\_FC}

To fairly compare with the Hybrid\_SRM\_FC (Fig.\ref{hybrid}), we further test the Hybrid\_LIF\_GNN (Fig.\ref{structures}) on ``Objects-v1'', ``Containers-v1'', and ``Slip Detection''. From Table \ref{comp}, we can see that the Hybrid\_LIF\_GNN outperforms the Hybrid\_SRM\_FC on ``Objects-v1'' and ``Containers-v1'' and they both achieve the perfect slip detection. The reason for this is that the Hybrid\_LIF\_GNN adopts graph topologies and has a more complicated structure than the Hybrid\_SRM\_FC. Such comparison results are consistent with the comparison between the Tactile-SNN and TactileSGNet in Table \ref{t2} and demonstrate the benefit of spiking graph neural networks and complex structures on event-driven tactile learning. Through this experiment, we show that the location spiking neurons can be incorporated into complex spike-based learning frameworks and further boost the performance of event-driven tactile learning.

\begin{table}[ht]\centering
	\begin{threeparttable}
		\caption{Performance comparison between the Hybrid\_SRM\_FC with LSRM neurons and the Hybrid\_LIF\_GNN with LLIF neurons}
		\setlength{\tabcolsep}{4mm}{
			\begin{tabular}{lcccc}
				\hline
				Method       & Type & Objects-v1       & Containers-v1    & Slip Detection \\ \hline
				Hybrid\_SRM\_FC  & SNN  & 0.91 & 0.86 & \textbf{1.0}   \\ \hline
				Hybrid\_LIF\_GNN$\ddagger$  & SNN  & \textbf{0.96} & \textbf{0.90} & \textbf{1.0}   \\ \hline
		\end{tabular}}\label{comp}
		\begin{tablenotes}
			\item $\ddagger$ represents Hybrid\_LIF\_GNN-sparse-mean-loop. The best performance is in bold.
		\end{tablenotes}
	\end{threeparttable}
\end{table}

\section{Discussion and Conclusion}\label{sec:conclude}
In this section, we discuss the advantages and limitations of conventional spiking neurons and location spiking neurons. Moreover, we provide preliminary results of the location spiking neurons on event-driven audio learning and discuss the potential impact of this work on broad spike-based learning applications. Finally, we conclude the paper.
\subsection{Advantages and Limitations of Conventional and Location Spiking Neurons}
This paper proposes location spiking neurons. \textbf{Based on the neuronal dynamic equations of conventional spiking neurons and location spiking neurons, we can see that both of them can extract spatio-temporal dependencies from the data.} Specifically, the conventional spiking neurons employ the \textbf{temporal} recurrent neural dynamics to update their membrane potentials and capture \textbf{spatial} dependencies by aggregating the information from presynaptic neurons, see Eqs. (\ref{e2}), (\ref{e7}), (\ref{e11}), and (\ref{e13}). However, location spiking neurons use \textbf{spatial} recurrent neural dynamics to update their potentials and model \textbf{temporal} dependencies by aggregating the information from presynaptic neurons, see Eqs. (\ref{e4}), (\ref{e10}), (\ref{e12}), and (\ref{e14}). 

Moreover, based on experimental results, we can see that conventional spiking neurons are better at capturing \textbf{spatial dependencies} which benefit the ``early'' accuracy, while location spiking neurons are better at modeling \textbf{mid-and-long temporal dependencies} which benefit the ``late'' accuracy. Networks built only with conventional spiking neurons or networks built only with location spiking neurons \textbf{cannot sufficiently} capture spatio-temporal dependencies in the event-based data. Thus, we always \textbf{concatenate or fuse} the networks to sufficiently capture spatio-temporal dependencies in the data.

By introducing LSRM neurons and LLIF neurons, we verify that the idea of location spiking neurons can be applied to various existing spiking neuron models like TSRM neurons and TLIF neurons and strengthen their feature representation abilities. Moreover, we extensively evaluate the models built with these novel neurons and demonstrate their superior performance and energy efficiency. Furthermore, by comparing the Hybrid\_LIF\_GNN with the Hybrid\_SRM\_FC, we show that the location spiking neurons can be utilized to build more complicated models to further improve task performance. 

\subsection{Potential Impact on Broad Spike-Based Learning Applications}
In this paper, we focus on boosting event-driven tactile learning with location spiking neurons. And extensive experimental results validate the effectiveness and efficiency of our models on the tasks. Besides event-driven tactile learning, we can also apply the models with location spiking neurons to other spike-based learning applications. To show the potential impact of our work, we apply the Hybrid\_SRM\_FC (see Fig. \ref{hybrid}) to event-driven audio learning and provide preliminary results. Please note that the objective of this experiment is not necessarily to obtain state-of-the-art results on event-driven audio learning, but to demonstrate that location spiking neurons can bring benefits to the model built with conventional spiking neurons on other spike-based learning applications. 

\begin{figure*}
	\includegraphics[width=\linewidth]{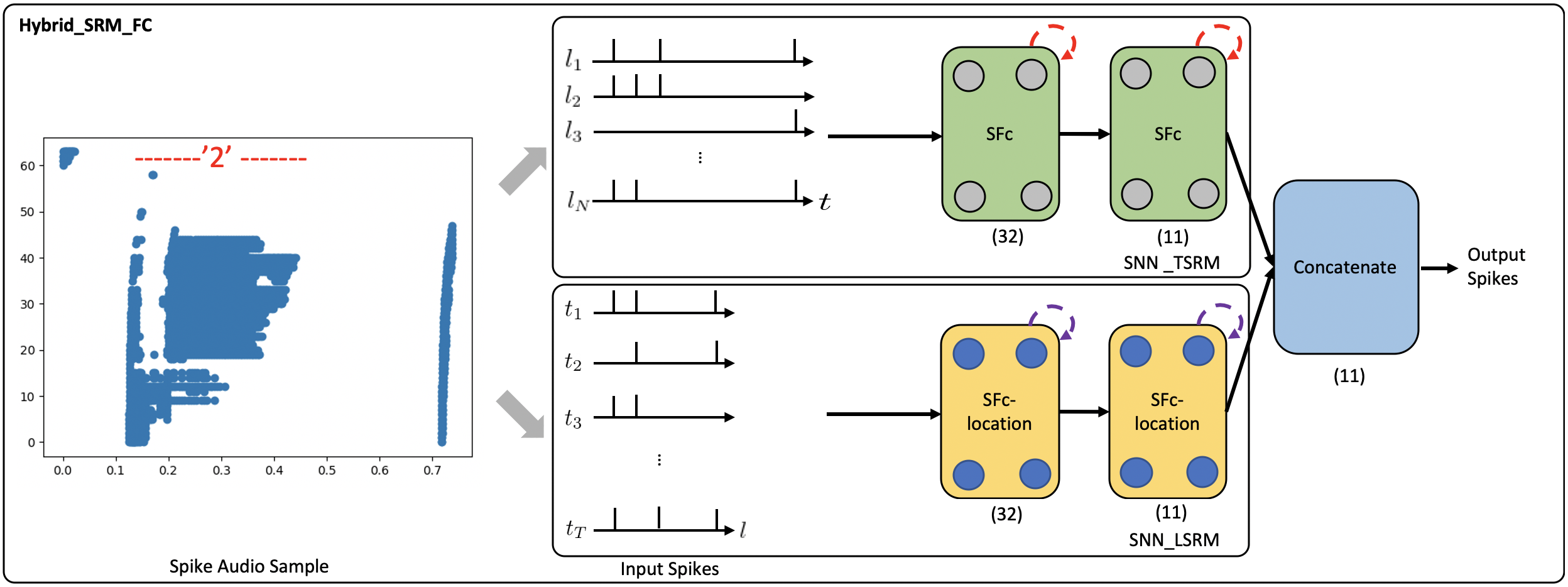}
	\caption{The Hybrid\_SRM\_FC processes a spike audio sequence and predict its label. The network structure of this model is the same as what we show in Fig. \ref{hybrid}. }
	\label{discuss_fig1}
\end{figure*}

In the experiment, we use the N-TIDIGITS18 dataset~\cite{anumula2018feature}, which is collected by playing the audio files from the TIDIGITS dataset~\cite{leonard1993tidigits} to the dynamic audio sensor -- the CochleaAMS1b sensor~\cite{chan2007aer}. The dataset includes both single digits and
connected digit sequences. We use the single-digit part of the dataset, which consists of 11 categories, including 'oh', 'zero', and digits '1' to '9'. 
A spike audio sequence of digit '2' is shown in Fig. \ref{discuss_fig1}, where the x-axis indicates the event time, and the y-axis indicates the 64 frequency channels of the CochleaAMS1b sensor. Each blue dot in the sequence represents an event that occurs at time $t_e$ and frequency $f_e$. In this application, we regard ``frequency channels'' as ``locations'' and apply the Hybrid\_SRM\_FC to process the spike audio inputs, see Fig. \ref{discuss_fig1}. Through the experiments, the fully-connected SNN with TSRM neurons achieves the test accuracy of 0.563. However, with the help of LSRM neurons, the Hybrid\_SRM\_FC obtains the test accuracy of 0.586 and \textbf{correctly classifies the additional 57 spike audio sequences}. Moreover, we show the training and testing profiles of the fully-connected SNN with TSRM neurons and the Hybrid\_SRM\_FC in the Supplementary Material. From those figures, we can see that our hybrid model converges faster and attains a lower loss and a higher accuracy compared to the fully-connected SNN with TSRM neurons. 

From this experiment, we can see that location spiking neurons can be applied to other spike-based learning applications. Moreover, the location spiking neurons can bring benefits to the models built with conventional spiking neurons and improve their task performance. We believe there will be further improvements on event-driven audio learning if we can incorporate the location spiking neurons into state-of-the-art event-driven audio learning frameworks.

Besides event-driven audio learning, a contemporary work~\cite{li2022brain} also validates the effectiveness of  spatial recurrent neuronal dynamics on conventional image classification. This work incorporates the spatial recurrent neuronal dynamics into the full-precision Multilayer Perceptron (MLP) and achieves the state-of-the-art top-1 accuracy on the ImageNet dataset. Since the model is full-precision and real-valued, it may lose the energy efficiency benefits of binary spikes. Our location spiking neurons employ the spatial recurrent neuronal dynamics but also keep the binary nature of spikes. Based on these, we think our proposed neurons could bring more potential to computer vision (e.g., event-based vision) when they are incorporated into MLP~\cite{tolstikhin2021mlp} or Transformer~\cite{dosovitskiy2020image} frameworks. 

\subsection{Conclusion}
In this work, we propose a novel neuron model -- ``location spiking neuron''. Specifically, we introduce two concrete location spiking neurons -- the LSRM neurons and LLIF neurons. We demonstrate the spatial recurrent neuronal dynamics of these neurons and compare them with the conventional spiking neurons -- the TSRM neurons and TLIF neurons. By exploiting these location spiking neurons, we develop two hybrid models for event-driven tactile learning to sufficiently capture the complex spatio-temporal dependencies in the event-based tactile data. The extensive experimental results on the event-driven tactile datasets demonstrate the extraordinary performance and high energy efficiency of our models and location spiking neurons. This could further unlock their potential on neuromorphic hardware. Overall, this work sheds new light on SNN representation learning and event-driven learning.


\section*{Acronyms and Notations}\label{Acronyms and Notations}
\captionsetup{singlelinecheck = false, justification=justified}
\begin{table}[ht]
	\caption{List of Acronyms and Notations in the Paper}
		\setlength{\tabcolsep}{0.3mm}{
			\begin{tabular}{p{33mm}l}
				
				\hline\vspace{-3mm}&\vspace{-3mm} \\
				TSRM & Time Spike Response Model \\
				LSRM & Location Spike Response Model \\
				TLIF & Time Leaky Integrate-and-Fire \\
				LLIF & Location Leaky Integrate-and-Fire \\
				Hybrid\_SRM\_FC & the hybrid model that combines a fully-connected SNN with TSRM neurons \\& and a fully-connected SNN with LSRM neurons \\
				Hybrid\_LIF\_GNN & the hybrid model that fuses the spatial spiking graph neural network with TLIF \\&neurons and temporal spiking graph neural network with LLIF neurons \\ 
				\hline\vspace{-3mm}&\vspace{-3mm} \\
				
				$\nu$ & $\nu=t$ for existing spiking neurons and $\nu=l$ for location spiking neurons \\
				$u_i(\nu)$  & the membrane potential of neuron $i$ at $\nu$   \\
				$\eta_i(\cdot)$ & the refractory kernel of neuron $i$ \\
				$\epsilon_{ij}(\cdot)$ & the incoming spike response kernel between neurons $i$ and $j$\\
				$\Gamma_i$ & the set of presynaptic neurons of neuron $i$\\
				$w_{ij}$ & the connection strength between neurons $i$ and $j$\\
				$x_j(\nu)$ & the presynaptic input from pre-neuron $j$ at $\nu$ \\
				$I(\nu)$  & the weighted summation of the presynaptic inputs at $\nu$  \\
				$\tau$  & the time constant of TLIF neurons    \\
				$\alpha$ & the decay factor of TLIF neurons \\
				$\tau'$  & the location constant of LLIF neurons    \\
				$\beta$ & the location decay factor of LLIF neurons \\
				$u_{th}$ & the firing thresholds of neurons \\ 
				\hline\vspace{-3mm}&\vspace{-3mm} \\
				
				
				
				
				$N$ & the number of taxels of NeuTouch \\ 
				$T$ & the number of total time length of event sequences \\
				$K$ & the number of classes for the tasks \\
				$X_{in}$ & the event-based tactile input \\
				$X'_{in}$ & the transposed event-based tactile input \\
				$O_1$ & output spikes from the SNN with existing spiking neurons \\
				$o_i(t)$ & the output spiking state of existing spiking neuron $i$ at time $t$ \\
				$O_2$ & output spikes from the SNN with location spiking neurons\\
				$o_i(l)$ & the output spiking state of location spiking neuron $i$ at location $l$ \\
				$O$ & output spikes from the Hybrid\_SRM\_FC \\ 
				
				$G_{s}(t)$ & the tactile spatial graph at time $t$ \\
				$G_t(n)$ & the tactile temporal graph of taxel $n$ \\
				$O_1'$ & the predicted label vector of the spatial spiking graph neural network\\
				$O_2'$ & the predicted label vector of the temporal spiking graph neural network\\
				$O'$ & the predicted label vector of the Hybrid\_LIF\_GNN\\
				\hline

	\end{tabular}}\label{notation_tab}
\end{table}

\section*{Acknowledgments}
Portions of this work ``Event-Driven Tactile Learning with Location Spiking Neurons~\cite{kangTactile}'' were accepted by IJCNN 2022 and orally presented at the IEEE WCCI in 2022.

\bibliography{sample}

\begin{thebibliography}{}

\bibitem[Abbott, 1999]{abbott1999lapicque}
Abbott, L.~F. (1999).
\newblock Lapicque’s introduction of the integrate-and-fire model neuron
  (1907).
\newblock {\em Brain research bulletin}, 50(5-6):303--304.

\bibitem[Anumula et~al., 2018]{anumula2018feature}
Anumula, J., Neil, D., Delbruck, T., and Liu, S.-C. (2018).
\newblock Feature representations for neuromorphic audio spike streams.
\newblock {\em Frontiers in neuroscience}, 12:23.

\bibitem[Baishya and B{\"a}uml, 2016]{baishya2016robust}
Baishya, S.~S. and B{\"a}uml, B. (2016).
\newblock Robust material classification with a tactile skin using deep
  learning.
\newblock In {\em 2016 IEEE/RSJ International Conference on Intelligent Robots
  and Systems (IROS)}, pages 8--15. IEEE.

\bibitem[Bullmore and Sporns, 2012]{bullmore2012economy}
Bullmore, E. and Sporns, O. (2012).
\newblock The economy of brain network organization.
\newblock {\em Nature Reviews Neuroscience}, 13(5):336--349.

\bibitem[Calandra et~al., 2018]{calandra2018more}
Calandra, R., Owens, A., Jayaraman, D., Lin, J., Yuan, W., Malik, J., Adelson,
  E.~H., and Levine, S. (2018).
\newblock More than a feeling: Learning to grasp and regrasp using vision and
  touch.
\newblock {\em IEEE Robotics and Automation Letters}, 3(4):3300--3307.

\bibitem[Cao et~al., 2015]{cao2015spiking}
Cao, Y., Chen, Y., and Khosla, D. (2015).
\newblock Spiking deep convolutional neural networks for energy-efficient
  object recognition.
\newblock {\em International Journal of Computer Vision}, 113(1):54--66.

\bibitem[Chan et~al., 2007]{chan2007aer}
Chan, V., Liu, S.-C., and van Schaik, A. (2007).
\newblock Aer ear: A matched silicon cochlea pair with address event
  representation interface.
\newblock {\em IEEE Transactions on Circuits and Systems I: Regular Papers},
  54(1):48--59.

\bibitem[Cheng et~al., 2020]{ijcai2020-211}
Cheng, X., Hao, Y., Xu, J., and Xu, B. (2020).
\newblock Lisnn: Improving spiking neural networks with lateral interactions
  for robust object recognition.
\newblock In Bessiere, C., editor, {\em Proceedings of the Twenty-Ninth
  International Joint Conference on Artificial Intelligence, {IJCAI-20}}, pages
  1519--1525. International Joint Conferences on Artificial Intelligence
  Organization.
\newblock Main track.

\bibitem[Cho et~al., 2014]{cho2014learning}
Cho, K., Van~Merri{\"e}nboer, B., Gulcehre, C., Bahdanau, D., Bougares, F.,
  Schwenk, H., and Bengio, Y. (2014).
\newblock Learning phrase representations using rnn encoder-decoder for
  statistical machine translation.
\newblock {\em arXiv preprint arXiv:1406.1078}.

\bibitem[Clevert et~al., 2015]{clevert2015fast}
Clevert, D.-A., Unterthiner, T., and Hochreiter, S. (2015).
\newblock Fast and accurate deep network learning by exponential linear units
  (elus).
\newblock {\em arXiv preprint arXiv:1511.07289}.

\bibitem[Davies et~al., 2021]{davies2021advancing}
Davies, M., Wild, A., Orchard, G., Sandamirskaya, Y., Guerra, G. A.~F., Joshi,
  P., Plank, P., and Risbud, S.~R. (2021).
\newblock Advancing neuromorphic computing with loihi: A survey of results and
  outlook.
\newblock {\em Proceedings of the IEEE}, 109(5):911--934.

\bibitem[Dosovitskiy et~al., 2020]{dosovitskiy2020image}
Dosovitskiy, A., Beyer, L., Kolesnikov, A., Weissenborn, D., Zhai, X.,
  Unterthiner, T., Dehghani, M., Minderer, M., Heigold, G., Gelly, S., et~al.
  (2020).
\newblock An image is worth 16x16 words: Transformers for image recognition at
  scale.
\newblock {\em arXiv preprint arXiv:2010.11929}.

\bibitem[Du et~al., 2017]{du2017topology}
Du, J., Zhang, S., Wu, G., Moura, J.~M., and Kar, S. (2017).
\newblock Topology adaptive graph convolutional networks.
\newblock {\em arXiv preprint arXiv:1710.10370}.

\bibitem[Felleman and Van~Essen, 1991]{felleman1991distributed}
Felleman, D.~J. and Van~Essen, D.~C. (1991).
\newblock Distributed hierarchical processing in the primate cerebral cortex.
\newblock In {\em Cereb cortex}. Citeseer.

\bibitem[Fishel and Loeb, 2012]{fishel2012sensing}
Fishel, J.~A. and Loeb, G.~E. (2012).
\newblock Sensing tactile microvibrations with the biotac—comparison with
  human sensitivity.
\newblock In {\em 2012 4th IEEE RAS \& EMBS international conference on
  biomedical robotics and biomechatronics (BioRob)}, pages 1122--1127. IEEE.

\bibitem[Gallego et~al., 2020]{gallego2020event}
Gallego, G., Delbruck, T., Orchard, G.~M., Bartolozzi, C., Taba, B., Censi, A.,
  Leutenegger, S., Davison, A., Conradt, J., Daniilidis, K., et~al. (2020).
\newblock Event-based vision: A survey.
\newblock {\em IEEE transactions on pattern analysis and machine intelligence}.

\bibitem[Gandarias et~al., 2019]{gandarias2019active}
Gandarias, J.~M., Pastor, F., Garc{\'\i}a-Cerezo, A.~J., and G{\'o}mez-de
  Gabriel, J.~M. (2019).
\newblock Active tactile recognition of deformable objects with 3d
  convolutional neural networks.
\newblock In {\em 2019 IEEE World Haptics Conference (WHC)}, pages 551--555.
  IEEE.

\bibitem[Gerstner, 1995]{gerstner1995time}
Gerstner, W. (1995).
\newblock Time structure of the activity in neural network models.
\newblock {\em Physical review E}, 51(1):738.

\bibitem[Gerstner and Kistler, 2002]{gerstner2002spiking}
Gerstner, W. and Kistler, W.~M. (2002).
\newblock {\em Spiking neuron models: Single neurons, populations, plasticity}.
\newblock Cambridge university press.

\bibitem[Gu et~al., 2020]{gu2020tactilesgnet}
Gu, F., Sng, W., Taunyazov, T., and Soh, H. (2020).
\newblock Tactilesgnet: A spiking graph neural network for event-based tactile
  object recognition.
\newblock In {\em 2020 IEEE/RSJ International Conference on Intelligent Robots
  and Systems (IROS)}, pages 9876--9882. IEEE.

\bibitem[Horowitz, 2014]{horowitz20141}
Horowitz, M. (2014).
\newblock 1.1 computing's energy problem (and what we can do about it).
\newblock In {\em 2014 IEEE International Solid-State Circuits Conference
  Digest of Technical Papers (ISSCC)}, pages 10--14. IEEE.

\bibitem[Kang et~al., 2022]{kangTactile}
Kang, P., Banerjee, S., Chopp, H., Katsaggelos, A., and Cossairt, O. (2022).
\newblock Event-driven tactile learning with location spiking neurons.
\newblock In {\em 2022 International Joint Conference on Neural Networks
  (IJCNN)}, pages 1--9. IEEE.

\bibitem[Kappassov et~al., 2015]{kappassov2015tactile}
Kappassov, Z., Corrales, J.-A., and Perdereau, V. (2015).
\newblock Tactile sensing in dexterous robot hands.
\newblock {\em Robotics and Autonomous Systems}, 74:195--220.

\bibitem[Lee et~al., 2020]{lee2020spike}
Lee, C., Kosta, A.~K., Zhu, A.~Z., Chaney, K., Daniilidis, K., and Roy, K.
  (2020).
\newblock Spike-flownet: event-based optical flow estimation with
  energy-efficient hybrid neural networks.
\newblock In {\em European Conference on Computer Vision}, pages 366--382.
  Springer.

\bibitem[Leonard and Doddington, 1993]{leonard1993tidigits}
Leonard, R.~G. and Doddington, G. (1993).
\newblock Tidigits speech corpus.
\newblock {\em Texas Instruments, Inc}.

\bibitem[Li et~al., 2016]{li2016evaluating}
Li, D., Chen, X., Becchi, M., and Zong, Z. (2016).
\newblock Evaluating the energy efficiency of deep convolutional neural
  networks on cpus and gpus.
\newblock In {\em 2016 IEEE international conferences on big data and cloud
  computing (BDCloud), social computing and networking (SocialCom), sustainable
  computing and communications (SustainCom)(BDCloud-SocialCom-SustainCom)},
  pages 477--484. IEEE.

\bibitem[Li et~al., 2022]{li2022brain}
Li, W., Chen, H., Guo, J., Zhang, Z., and Wang, Y. (2022).
\newblock Brain-inspired multilayer perceptron with spiking neurons.
\newblock In {\em Proceedings of the IEEE/CVF Conference on Computer Vision and
  Pattern Recognition}, pages 783--793.

\bibitem[Maas et~al., 2013]{maas2013rectifier}
Maas, A.~L., Hannun, A.~Y., Ng, A.~Y., et~al. (2013).
\newblock Rectifier nonlinearities improve neural network acoustic models.
\newblock In {\em Proc. icml}, volume~30, page~3. Citeseer.

\bibitem[Maass and Bishop, 2001]{maass2001pulsed}
Maass, W. and Bishop, C.~M. (2001).
\newblock {\em Pulsed neural networks}.
\newblock MIT press.

\bibitem[Merolla et~al., 2014]{merolla2014million}
Merolla, P.~A., Arthur, J.~V., Alvarez-Icaza, R., Cassidy, A.~S., Sawada, J.,
  Akopyan, F., Jackson, B.~L., Imam, N., Guo, C., Nakamura, Y., et~al. (2014).
\newblock A million spiking-neuron integrated circuit with a scalable
  communication network and interface.
\newblock {\em Science}, 345(6197):668--673.

\bibitem[Pfeiffer and Pfeil, 2018]{pfeiffer2018deep}
Pfeiffer, M. and Pfeil, T. (2018).
\newblock Deep learning with spiking neurons: opportunities and challenges.
\newblock {\em Frontiers in neuroscience}, 12:774.

\bibitem[Roy et~al., 2019]{roy2019towards}
Roy, K., Jaiswal, A., and Panda, P. (2019).
\newblock Towards spike-based machine intelligence with neuromorphic computing.
\newblock {\em Nature}, 575(7784):607--617.

\bibitem[Sanchez et~al., 2018]{sanchez2018online}
Sanchez, J., Mateo, C.~M., Corrales, J.~A., Bouzgarrou, B.-C., and Mezouar, Y.
  (2018).
\newblock Online shape estimation based on tactile sensing and deformation
  modeling for robot manipulation.
\newblock In {\em 2018 IEEE/RSJ International Conference on Intelligent Robots
  and Systems (IROS)}, pages 504--511. IEEE.

\bibitem[Schmitz et~al., 2010]{schmitz2010tactile}
Schmitz, A., Maggiali, M., Natale, L., Bonino, B., and Metta, G. (2010).
\newblock A tactile sensor for the fingertips of the humanoid robot icub.
\newblock In {\em 2010 IEEE/RSJ International Conference on Intelligent Robots
  and Systems}, pages 2212--2217. IEEE.

\bibitem[Sengupta et~al., 2019]{sengupta2019going}
Sengupta, A., Ye, Y., Wang, R., Liu, C., and Roy, K. (2019).
\newblock Going deeper in spiking neural networks: Vgg and residual
  architectures.
\newblock {\em Frontiers in neuroscience}, 13:95.

\bibitem[Shrestha and Orchard, 2018]{Shrestha2018}
Shrestha, S.~B. and Orchard, G. (2018).
\newblock {SLAYER}: Spike layer error reassignment in time.
\newblock In Bengio, S., Wallach, H., Larochelle, H., Grauman, K.,
  Cesa-Bianchi, N., and Garnett, R., editors, {\em Advances in Neural
  Information Processing Systems 31}, pages 1419--1428. Curran Associates, Inc.

\bibitem[Soh and Demiris, 2014]{soh2014incrementally}
Soh, H. and Demiris, Y. (2014).
\newblock Incrementally learning objects by touch: Online discriminative and
  generative models for tactile-based recognition.
\newblock {\em IEEE transactions on haptics}, 7(4):512--525.

\bibitem[Strubell et~al., 2019]{strubell2019energy}
Strubell, E., Ganesh, A., and McCallum, A. (2019).
\newblock Energy and policy considerations for deep learning in nlp.
\newblock {\em arXiv preprint arXiv:1906.02243}.

\bibitem[Taunyazov et~al., 2020]{taunyazov2020fast}
Taunyazov, T., Chua, Y., Gao, R., Soh, H., and Wu, Y. (2020).
\newblock Fast texture classification using tactile neural coding and spiking
  neural network.
\newblock In {\em 2020 IEEE/RSJ International Conference on Intelligent Robots
  and Systems (IROS)}, pages 9890--9895. IEEE.

\bibitem[Taunyazov et~al., 2019]{taunyazov2019towards}
Taunyazov, T., Koh, H.~F., Wu, Y., Cai, C., and Soh, H. (2019).
\newblock Towards effective tactile identification of textures using a hybrid
  touch approach.
\newblock In {\em 2019 International Conference on Robotics and Automation
  (ICRA)}, pages 4269--4275. IEEE.

\bibitem[Taunyazov et~al., 2021]{taunyazov2021extended}
Taunyazov, T., Song, L.~S., Lim, E., See, H.~H., Lee, D., Tee, B.~C., and Soh,
  H. (2021).
\newblock Extended tactile perception: Vibration sensing through tools and
  grasped objects.
\newblock {\em arXiv preprint arXiv:2106.00489}.

\bibitem[Taunyazoz et~al., 2020]{taunyazov2020event}
Taunyazoz, T., Sng, W., See, H.~H., Lim, B., Kuan, J., Ansari, A.~F., Tee, B.,
  and Soh, H. (2020).
\newblock Event-driven visual-tactile sensing and learning for robots.
\newblock In {\em Proceedings of Robotics: Science and Systems}.

\bibitem[Tolstikhin et~al., 2021]{tolstikhin2021mlp}
Tolstikhin, I.~O., Houlsby, N., Kolesnikov, A., Beyer, L., Zhai, X.,
  Unterthiner, T., Yung, J., Steiner, A., Keysers, D., Uszkoreit, J., et~al.
  (2021).
\newblock Mlp-mixer: An all-mlp architecture for vision.
\newblock {\em Advances in Neural Information Processing Systems},
  34:24261--24272.

\bibitem[Wu et~al., 2018]{wu2018spatio}
Wu, Y., Deng, L., Li, G., Zhu, J., and Shi, L. (2018).
\newblock Spatio-temporal backpropagation for training high-performance spiking
  neural networks.
\newblock {\em Frontiers in neuroscience}, 12:331.

\bibitem[Xu et~al., 2015]{xu2015empirical}
Xu, B., Wang, N., Chen, T., and Li, M. (2015).
\newblock Empirical evaluation of rectified activations in convolutional
  network.
\newblock {\em arXiv preprint arXiv:1505.00853}.

\bibitem[Xu et~al., 2021]{ijcai2021-441}
Xu, M., Wu, Y., Deng, L., Liu, F., Li, G., and Pei, J. (2021).
\newblock Exploiting spiking dynamics with spatial-temporal feature
  normalization in graph learning.
\newblock In Zhou, Z.-H., editor, {\em Proceedings of the Thirtieth
  International Joint Conference on Artificial Intelligence, {IJCAI-21}}, pages
  3207--3213. International Joint Conferences on Artificial Intelligence
  Organization.
\newblock Main Track.

\end{thebibliography}

\end{document}